\documentclass{article}

\usepackage[final]{neurips_2021}

\usepackage[utf8]{inputenc} 
\usepackage[T1]{fontenc}    
\usepackage{hyperref}       
\usepackage{url}            
\usepackage{booktabs}       
\usepackage{amsfonts}       
\usepackage{nicefrac}       
\usepackage{microtype}      
\usepackage{xcolor}         

\usepackage{enumitem}
\usepackage{bm} 
\usepackage{caption}
\usepackage[normalem]{ulem}
\usepackage{algorithmic}
\usepackage{amsmath,amssymb,amsthm}

\usepackage{microtype}
\usepackage{graphicx}
\usepackage{subfigure}

\newcommand{\swaptext}[2]{#2} 

\newcommand{\ench}[1]{\mathbf{h}^\text{E}_{#1}}
\newcommand{\dech}[1]{\mathbf{h}^\text{D}_{#1}}
\newcommand{\enct}[1]{\bm{\mu}^\text{E}_{#1}}
\newcommand{\dect}[1]{\bm{\mu}^\text{D}_{#1}}
\newcommand{\enci}[1]{\bm{\chi}^\text{E}_{#1}}
\newcommand{\deci}[1]{\bm{\chi}^\text{D}_{#1}}
\newcommand{\encd}[1]{\Delta\mathbf{h}^\text{E}_{#1}}
\newcommand{\decd}[1]{\Delta\mathbf{h}^\text{D}_{#1}}

\title{Understanding How Encoder-Decoder \\ Architectures Attend}

\author{%
  Kyle Aitken \\
  Department of Physics\\
  University of Washington\\
  Seattle, Washington, USA \\
  \texttt{kaitken17@gmail.com} \\
  \And
  Vinay V Ramasesh \\
  Google Research, Blueshift Team \\
  Mountain View, California, USA \\
  \AND
  Yuan Cao \\
  Google Research, Brain Team \\
  Mountain View, California, USA \\
  \And
  Niru Maheswaranathan \\
  Google Research, Brain Team \\
  Mountain View, California, USA \\
}

\begin{document}

\maketitle

\begin{abstract}
    Encoder-decoder networks with attention have proven to be a powerful way to solve many sequence-to-sequence tasks. 
    In these networks, attention aligns encoder and decoder states and is often used for visualizing network behavior.
    However, the mechanisms used by networks to generate appropriate attention matrices are still mysterious.
    Moreover, how these mechanisms vary depending on the particular architecture used for the encoder and decoder (recurrent, feed-forward, etc.) are also not well understood.
    In this work, we investigate how encoder-decoder networks solve different sequence-to-sequence tasks.
    We introduce a way of decomposing hidden states over a sequence into \textit{temporal} (independent of input) and \textit{input-driven} (independent of sequence position) components.
    This reveals how attention matrices are formed: depending on the task requirements, networks rely more heavily on either the temporal or input-driven components.
    These findings hold across both recurrent and feed-forward architectures despite their differences in forming the temporal components.
    Overall, our results provide new insight into the inner workings of attention-based encoder-decoder networks.
\end{abstract}

\section{Introduction}

Modern machine learning encoder-decoder architectures can achieve strong performance on sequence-to-sequence tasks such as machine translation \citep{bahdanau2014neural, luong2015effective, wu2016googles, vaswani2017attention}, language modeling \citep{raffel2020exploring}, speech-to-text \citep{chan2015listen, rohit2017comparison, chiu2018state}, etc.
Many of these architectures make use of attention \citep{bahdanau2014neural}, a mechanism that allows the network to focus on a specific part of the input most relevant to the current prediction step.
Attention has proven to be a critical mechanism; indeed many modern architectures, such as the Transformer, are fully attention-based \citep{vaswani2017attention}.
However, despite the success of these architectures, an understanding of \emph{how} said networks solve such tasks using attention remains largely unknown.

Attention mechanisms are attractive because they are interpretable, and often illuminate key computations required for a task.
For example, consider neural machine translation---trained networks exhibit attention matrices that align words in the encoder sequence with the appropriate corresponding position in the decoder sentence~\citep{ghader2017does,ding2019saliency}.
In this case, the attention matrix already contains information about which words in the source sequence are relevant for translating a particular word in the target sequence; that is, forming the attention matrix itself constitutes a significant part of solving the overall task.
How is it that networks are able to achieve this?
What are the mechanisms underlying how networks form attention, and how do they vary across tasks and architectures?

In this work, we study these questions by analyzing three different encoder-decoder architectures on sequence-to-sequence tasks.
We develop a method for decomposing the hidden states of the network into a sum of components that let us isolate \emph{input} driven behavior from \emph{temporal} (or sequence) driven behavior.
We use this to first understand how networks solve tasks where all samples use the same attention matrix, a diagonal one. 
We then build on that to show how additional mechanisms can generate sample-dependent attention matrices that are still close to the average matrix.

\textbf{Our Contributions}
\begin{itemize}[leftmargin=0.2in] 
\itemsep-0.0em 
    \item We propose a decomposition of hidden state dynamics into  separate pieces, one of which explains the temporal behavior of the network, another of which describes the input behavior. 
    We show such a decomposition aids in understanding the behavior of networks with attention.
    \item In the tasks studied, we show the temporal (input) components play a larger role in determining the attention matrix as the average attention matrix becomes a better (worse) approximation for a random sample’s attention matrix.
    \item We discuss the dynamics of architectures with attention and/or recurrence and show how the input/temporal component behavior differs across said architectures.
    \item We investigate the detailed temporal and input component dynamics in a synthetic setting to understand the mechanism behind common sequence-to-sequence structures and how they might differ in the presence of recurrence.
\end{itemize}

\textbf{Related Work} \hspace{1mm}
As mentioned in the introduction, a common technique to gain some understanding is to visualize learned attention matrices, though the degree to which such visualization can explain model predictions is disputed~\cite{wiegreffe2019attention,jain2019attention, serrano2019attention}.  Input saliency~\cite{bastings2020elephant} and attribution-propagation~\cite{chefer2020transformer} methods have also been studied as potential tools for model interpretability.

Complementary to these works, our approach builds on a recent line of work analyzing the computational mechanisms learned by RNNs from a dynamical systems perspective. 
These analyses have identified simple and interpretable hidden state dynamics underlying RNN operation on text-classification tasks such as binary sentiment analysis \citep{Maheswaranathan2019, maheswaranathan2020recurrent} and document classification~\citep{aitken2020geometry}. 
Our work extends these ideas into the domain of sequence-to-sequence tasks.

\textbf{Notation} \hspace{0.5mm}
Let $T$ and $S$ be the input and output sequence length of a given sample, respectively.
We denote the encoder and decoder hidden states by $\ench{t} \in \mathbb{R}^n$ with $t=1,\ldots,T$. 
Similarly, we denote decoder hidden states by $\dech{s} \in \mathbb{R}^n$, with $s=1,\ldots,S$.
The encoder and decoder hidden state dimensions are always taken to be equal in this work.
Inputs to the encoder and decoder are denoted by $\mathbf{x}_t^\text{E} \in \mathbb{R}^d$ and $\mathbf{x}_s^\text{D} \in \mathbb{R}^{\tilde{d}}$. 
When necessary, we subscript different samples from a test/train set using $\alpha, \beta, \gamma$, e.g. $\mathbf{x}_{t,\alpha}^\text{E}$ for $\alpha = 1,\ldots, M$.

\textbf{Outline} \hspace{0.5mm}
We begin by introducing the three architectures we investigate in this work with varying combinations of recursion and attention. 
Next we introduce our temporal and input component decomposition and follow this up with a demonstration of how such a decomposition allows us to understand the dynamics of attention in a simple one-to-one translation task.
Afterwards, we apply this decomposition to two additional tasks with increasing levels of complexity and discuss how our decomposition gives insight into the behavior of attention in these tasks.

\section{Setup}

\begin{figure}[t]
    \centering
    \includegraphics[width=0.7\textwidth]{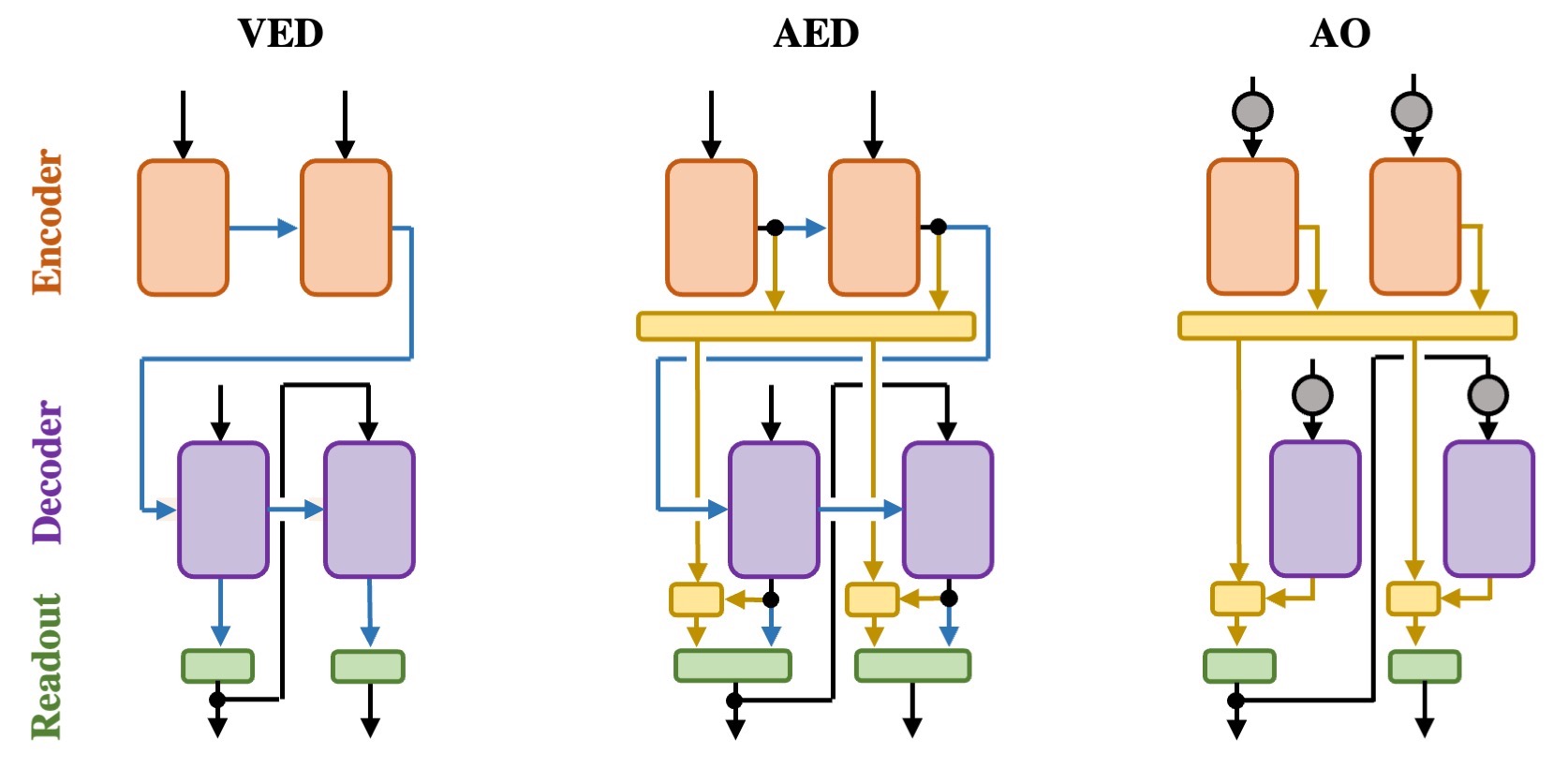}
    \vspace{-1mm}
    \caption{\small
    \textbf{Schematic of the three primary architectures analyzed in this work.} The orange, purple, and green boxes represent the encoder RNNs, decoder RNNs, and linear readout layers, respectively. 
    Recurrent connections are shown in blue, attention-based connections and computational blocks are shown in gold.
    The grey circles add positional encoding to the inputs.
    }
    \label{fig:arch_small}
    \vspace{-4mm}
\end{figure}

A schematic of the three architectures we study is shown in Fig.~\ref{fig:arch_small} (see \swaptext{SM}{Appendix \ref{app:archs}} for precise expressions).

\textbf{Vanilla Encoder Decoder (VED)} is a recurrent encoder-decoder architecture with no attention \citep{sutskever2014sequence}. 
The encoder and decoder update expression are $\ench{t} = F_\text{E}(\ench{t-1}, \mathbf{x}_t^\text{E})$ and $\dech{s} = F_\text{D}(\dech{s-1},\mathbf{x}_s^\text{D})$, respectively. 
Here, $F_\text{D}$ and $F_\text{E}$ are functions that implement the hidden state updates, which in this work are each one of three modern RNN cells: LSTMs~\citep{HochSchm97}, GRUs~\citep{GRU}, or UGRNNs~\citep{UGRNN}.

\textbf{Encoder-Decoder with Attention (AED)} is identical to the VED architecture above with a simple attention mechanism added \citep{bahdanau2014neural, luong2015effective}.
For time step $s$ of the decoder, we compute a  context vector $\mathbf{c}_s$, a weighted sum of encoder hidden states,
$\mathbf{c}_s := \sum_{t=1}^T \alpha_{st} \ench{t}$, with $\bm{\alpha}_{t} := \text{softmax}\left(a_{1t}, \ldots, a_{St} \right)$ the $t^\text{th}$ column of the \emph{attention matrix} and $a_{st}:=\dech{s}\cdot \ench{t}$ the \emph{alignment} between a given decoder and encoder hidden state. 
While more complicated attention mechanisms exist, in the main text we analyze the simplest form of attention for convenience of analysis.\footnote{
In \swaptext{the SM}{Appendix \ref{app:learned_attention}}, we implement a learned-attention mechanism using a scaled-dot product attention in the form of queries, keys, and value matrices \citep{vaswani2017attention}. For the AED and AO architectures, we find qualitatively similar results to the simple dot-product attention presented in the main text.}

\textbf{Attention Only (AO)} is identical to the AED network above, but simply eliminates the recurrent information passed from one RNN cell to the next and instead adds fixed positional encoding vectors  to the encoder and decoder inputs \citep{vaswani2017attention}.
Due to the lack of recurrence, the RNN functions $F_\text{E}$ and $F_\text{D}$ simply act as feedforward networks in this setting.\footnote{We train non-gated feedforward networks and find their dynamics to be qualitatively the same, see \swaptext{SM}{Appendix \ref{app:more_results}}.}
AO can be treated as a simplified version of a Transformer without self-attention, hence our analysis may also provide a hint into their inner workings \citep{vaswani2017attention}.

\subsection{Temporal and Input Components}

In architectures with attention, we will show that it is helpful to write the hidden states using what we will refer to as their \emph{temporal} and \emph{input} components.
This will be useful because each hidden state has an associated time step and input word at that same time step (e.g. $s$ and $\mathbf{x}_s^\text{D}$ for $\dech{s}$), therefore such a decomposition will often allow us to disentangle temporal and input behavior from any other network dynamics.

We define the temporal components of the encoder and decoder to be the average hidden state at a given time step, which we denote by $\enct{t}$ and $\dect{s}$, respectively.
Similarly, we define an encoder input component to be the average of all $\ench{t} - \enct{t}$ for hidden states that immediately follow a given input word.
We analogously define the decoder input components. 
In practice, we estimate such averages using a test set of size $M$, so that the temporal and input components of the encoder are respectively given by
\begin{align}
\enct{t} \approx & \frac{\sum_{
\alpha=1}^M \mathbf{1}_{\leq\text{EoS}, \alpha} \ench{t, \alpha}}{\sum_{
\beta=1}^M \mathbf{1}_{\leq\text{EoS}, \beta}}\, , \qquad
\bm{\chi}^\text{E}\left(\mathbf{x}_{t, \alpha}\right) \approx \frac{\sum_{\beta=1}^M \sum_{t^\prime=1}^T \mathbf{1}_{\mathbf{x}_{t,\alpha}, \mathbf{x}_{t^\prime,\beta}} \left(\ench{t^\prime,\beta} - \enct{t^\prime}\right)}{\sum_{\gamma=1}^M \sum_{t^{\prime\prime}=1}^T \mathbf{1}_{\mathbf{x}_{t,\alpha}, \mathbf{x}_{t^{\prime\prime},\gamma}}}\,,
\end{align}
where $\ench{t, \alpha} $ the encoder hidden state of the $\alpha$th sample, $\mathbf{1}_{\leq\text{EoS}, \alpha}$ is a mask that is zero if the $\alpha$th sample is beyond the end of sentence, $\mathbf{1}_{\mathbf{x}_{t,\alpha}, \mathbf{x}_{t^\prime,\beta}}$ is a mask that is zero if $\mathbf{x}_{t,\alpha} \ne \mathbf{x}_{t^\prime,\beta}$, and we have temporarily suppressed superscripts on the inputs for brevity.\footnote{
See \swaptext{SM}{Appendix \ref{app:methods}} for more details on this definition and the analogous decoder definitions.}
By definition, \emph{the temporal components only vary with time and the input components only vary with input/output word}.
As such, it will be useful to denote the encoder and decoder input components by $\enci{x}$ and $\deci{y}$, with $x$ and $y$ respectively running over all input and output words (e.g. $\enci{\text{yes}}$ and $\deci{\text{oui}}$).
We can then write any hidden state as
\begin{align}
\label{eq:hs_temporal_word}
    \ench{t} = \enct{t} + \enci{x} + \Delta\ench{t}  \, , \quad \dech{s} = \dect{s} + \deci{y} + \Delta\dech{s} \, ,
\end{align}
with $\Delta\ench{t} := \ench{t}- \enct{t} - \enci{t}$ and $\Delta\dech{s} := \dech{s} - \dect{s} - \deci{y}$ the \emph{delta components} of encoder and decoder hidden states, respectively.
Intuitively, we are simply decomposing each hidden state vector as a sum of a component that only varies with time/position in the sequence (independent of input), a component that only varies with input (independent of position), and whatever else is left over. 
Finally, we will often refer to hidden states without their temporal component, i.e. $\enci{x}+\encd{t}$ and $\deci{y}+\decd{s}$, so for brevity we refer to these combinations as the \emph{input-delta components}.

Using the temporal and input components in \eqref{eq:hs_temporal_word}, we can decompose the attention alignment between two hidden states as
\begin{align}
\label{eq:alignment}
    a_{st} = \left(\dect{s} + \deci{y} + \Delta\dech{s}\right) \cdot \left(\enct{t} + \enci{x} + \Delta\ench{t}\right)\,. 
\end{align}
We will show below that in certain cases several of the nine terms of this expression approximately vanish, leading to simple and interpretable attention mechanisms.

\begin{figure*}[t!]
    \centering
    \includegraphics[width=0.95\textwidth]{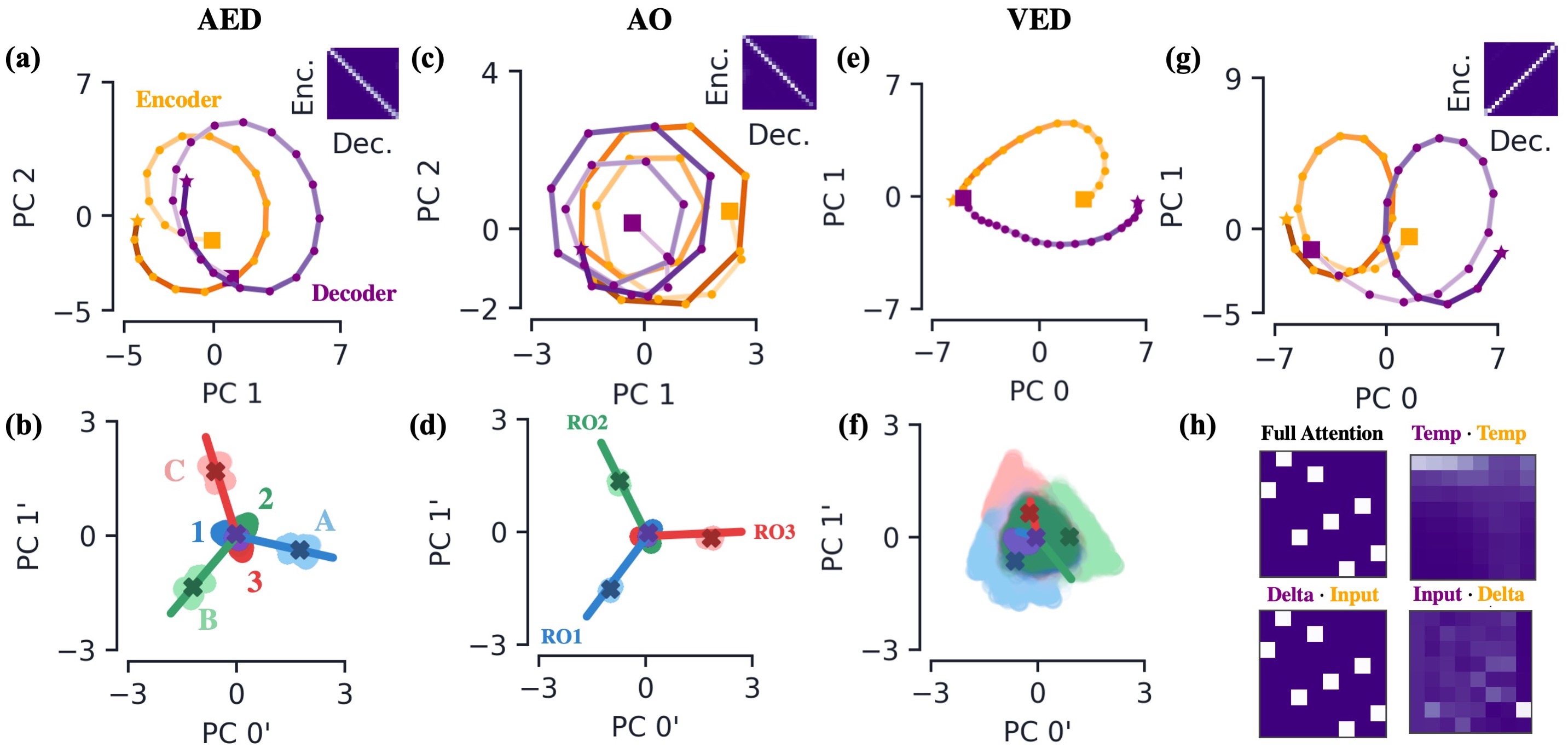}
    \vspace{-1mm}
    \caption{\small
    \textbf{Summary of attention dynamics on synthetic tasks.} 
    \textbf{(a-f)} All three architectures trained on an $N=3$ one-to-one translation task of variable length ranging from $15$ to $20$.
    Plots in the top row are projected onto the principal components (PCs) of the encoder and decoder temporal components, while those in the bottom row are projected onto the PCs of the input components.
    \textbf{(a)} For AED, the path formed by the temporal components of the encoder (orange) and decoder (purple), $\enct{t}$ and $\dect{s}$.
    We denote the first and last temporal component by a square and star, respectively, and the color of said path is lighter for earlier times.
    The inset shows the softmaxed alignment scores for $\dect{s} \cdot \enct{t}$, which we find to be a good approximation to the full alignment for the one-to-one translation task.
    \textbf{(b)} The input-delta components of the encoder (light) and decoder (dark) colored by word (see labels).
    The encoder input components, $\enci{x}$ are represented by a dark colored `X'.
    The solid lines are the readout vectors (see labels on (d)).
    Start/end of sentence characters are in purple.
    \textbf{(c, d)} The same plots for the AO network.
    \textbf{(e, f)} The same plots for the VED network (with no attention inset).
    \textbf{(g)} Temporal components for the same task with a temporally reversed output sequence.
    \textbf{(h)}  Attention matrices for a test example from a network trained to alphabetically sort a list of letters.
    Clockwise from top left, the softmaxed attention from the full hidden states ($\dech{s}\cdot\ench{t}$), temporal components only ($\dect{s}\cdot\enct{t}$), decoder input components and encoder delta components ($\deci{y}\cdot\encd{t}$), and decoder delta components and encoder input components ($\decd{s}\cdot\enci{x}$).
    }
    \label{fig:summary}
    \vspace{-4mm}
\end{figure*}

\section{One-to-One Results}
To first establish a basis of how each of the three architectures learn to solve tasks and the role of their input and temporal components, we start by studying their dynamics for a synthetic one-to-one translation task. 
The task is to convert a sequence of input words into a corresponding sequence of output words, where there is a one-to-one translation dictionary, e.g. converting a sequence of letters to their corresponding position in the alphabet, $\{\text{B}, \text{A}, \text{C}, \text{A}, \text{D}\} \to \{2, 1, 3, 1, 4\}$.
We generate the input phrases to have variable length, but outputs always have equal length to their input (i.e. $T=S$).
While a solution to this task is trivial, it is not obvious how each neural network architecture will solve the task.
Although this is a severely simplified approximation to realistic sequence-to-sequence tasks, we will show below that many of the dynamics the AED and AO networks learn on this task are qualitatively present in more complex tasks.

\textbf{Encoder-Decoder with Attention.} \hspace{0.5mm}
After training the AED architecture, we apply the decomposition of \eqref{eq:hs_temporal_word} to the hidden states.
Plotting the temporal components of both the encoder and decoder, they each form an approximate circle that is traversed as their respective inputs are read in (Fig.~\ref{fig:summary}a).\footnote{
Here and in plots that follow, we plot the various components using principal component analysis (PCA) projections simply as a convenient visualization tool. 
Other than observation that in some cases the temporal/input components live in a low-dimensional subspace, none of our quantitative analysis is dependent upon the PCA projections. 
For all one-to-one plots, a large percentage ($>90$\%) of the variance is explained by the first 2 or 3 PC dimensions. 
}
Additionally, we find the encoder and decoder temporal components are closest to alignment when $s=t$.
We also plot the input components of the encoder and decoder together with the encoder input-delta components, i.e. $\enci{x}+\encd{t}$, and the network's readout vectors (Fig.~\ref{fig:summary}b).\footnote{
For $N$ possible input words, the encoder input components align with the vetrices of an $(N-1)$-simplex, which is similar to the classification behavior observed in \citet{aitken2020geometry}. 
}
We see for the encoder hidden states, the input-delta components are clustered close to their respective input components, meaning for this task the delta components are negligible.
Also note the decoder input-delta components are significantly smaller in magnitude than the decoder temporal components.
Together, this means we can approximate the encoder and decoder hidden states as $\ench{t}\approx \enct{t}+\enci{x}$ and $\dech{s}\approx \dect{s}$, respectively.
Finally, note the readout vector for a given output word aligns with the input components of its translated input word, e.g. the readout for `$1$' aligns with the input component for `$\text{A}$' (Fig.~\ref{fig:summary}b).\footnote{Since in AED we pass both the decoder hidden state and the context vector to the readout, each readout vector is twice the hidden state dimension. 
We plot only the readout weights corresponding to the context vector, since generally those corresponding to the decoder hidden state are negligible, see \swaptext{SM}{Appendix~\ref{app:more_results}} for more details.}

For the one-to-one translation task, the network learns an approximately diagonal attention matrix, meaning the decoder at time $s$ primarily attends to the encoder's hidden state at $t=s$.
Additionally, we find the temporal and input-delta components to be close to orthogonal for all time steps, which allows the network's attention mechanism to isolate temporal dependence rather than input dependence.
Since we can approximate the hidden states as $\ench{t}\approx \enct{t}+\enci{x}$ and $\dech{s}\approx \dect{s}$, and the temporal and input components are orthogonal, the alignment in \eqref{eq:alignment} can be written simply as $a_{st}\approx \dect{s}\cdot\enct{t}$.
This means that the \emph{full} attention is completely described by the temporal components and thus input-independent (this will not necessarily be true for other tasks, as we will see later).

With the above results, we can understand how AED solves the one-to-one translation task. 
After reading a given input, the encoder hidden state is primarily composed of an input and temporal component that are approximately orthogonal to one another, with the input component aligned with the readout of the translated input word (Fig.~\ref{fig:summary}b). 
The decoder hidden states are approximately made up of only a temporal component, whose sole job is to align with the corresponding encoder temporal component.
Temporal components of the decoder and encoder are closest to alignment for $t=s$, so the network primarily attends to the encoder state $\ench{t=s}$. 
The alignment between encoder input components and readouts yields maximum logit values for the correct translation.

\textbf{Attention Only.} \hspace{0.5mm}
Now we turn to AO architecture, which is identical to AED except with the recurrent connections cut, and positional encoding added to the inputs.
We find that AO has qualitatively similar temporal components that give rise to diagonal attention~(Fig.~\ref{fig:summary}c) and the input components align with the readouts (Fig.~\ref{fig:summary}d).
Thus AO solves the task in a similar manner as AED.
The only difference is that the temporal components, driven by RNN dynamics in AED, are now driven purely by the positional encoding in AO.



\textbf{Vanilla Encoder-Decoder.} \hspace{0.5mm}
After training the VED architecture, we find the encoder and decoder hidden states belonging to the same time step form clusters, and said clusters are closest to those corresponding to adjacent time steps.
This yields temporal components that are close to one another for adjacent times, with $\enct{T}$ next to $\dect{1}$ (Figs.~\ref{fig:summary}e).
Since there is no attention in this architecture, there is no incentive for the network to align temporal components of the encoder and decoder as we saw in AED and AO.

\begin{figure*}[t]
    \centering
    \includegraphics[width=0.95\textwidth]{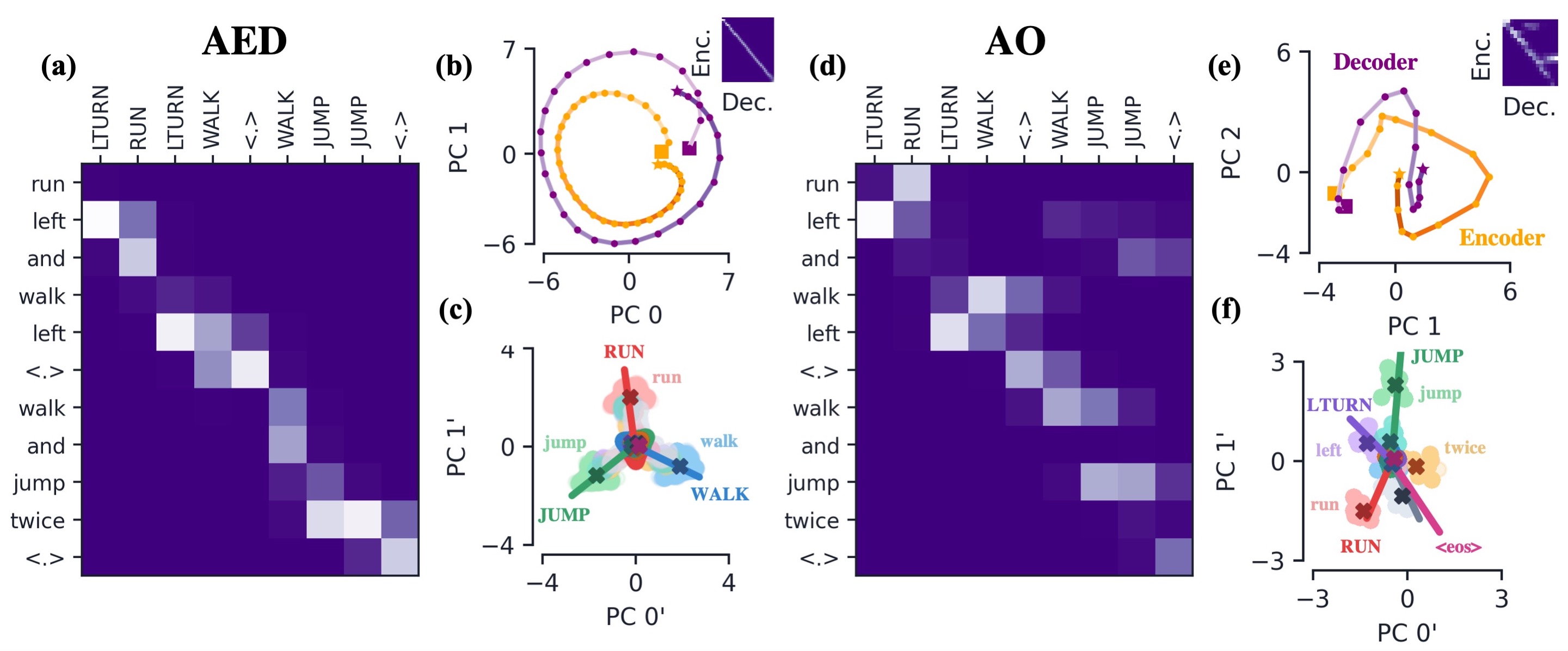}
    \vspace{-1mm}
    \caption{\small
    \textbf{Summary of dynamics for AED and AO architectures trained on eSCAN.}
    \textbf{(a)} Example attention matrix for the AED architecture.
    \textbf{(b)} AED network's temporal components, with the inset showing the attention matrix from said temporal components. 
    Once again, encoder and decoder components are orange and purple, respectively and we are projecting onto the temporal component PCs.
    \textbf{(c)} AED network's input-delta components, input components, and readouts, all colored by their corresponding input/output words (see labels).
    All quantities projected onto input component PCs.
    \textbf{(d, e, f)} The same plots for AO.
    }
    \label{fig:scan_qual}
    \vspace{-4mm}
\end{figure*}

As recurrence is the only method of transferring information across time steps, encoder and decoder hidden states must carry all relevant information from preceding steps.
Together, this results in the delta components deviating significantly more from their respective input components for VED relative to AED and AO (Fig.~\ref{fig:summary}f).
That is, since hidden states must hold the information of inputs/outputs for \emph{multiple} time steps, we cannot expect them to be well approximated by $\enct{t} + \enci{x}$ because, by definition, it is agnostic to the network's inputs at any time other than $t$ (and similarly for $\enct{s} + \enci{y}$).
As such, the temporal and input component decomposition gains us little insight into the inner workings of the VED architecture.
Additional details of the VED architecture dynamics are discussed in \swaptext{the SM}{Appendix~\ref{app:ved_dyn}}.

\textbf{Additional Tasks.} \hspace{0.5mm}
In this section, we briefly address how two additional synthetic tasks can be understood using the temporal and input component decomposition.
First, consider a task identical to the one-to-one task, with the target sequence reversed in time, e.g. $\{\text{B}, \text{A}, \text{C}, \text{A}, \text{D}\} \to \{4, 1, 3, 1, 2\}$.
For this task, we expect an attention matrix that is anti-diagonal (i.e. it is nonzero for $t = S + 1 - s$).
For the AED and AO networks trained on this task, we find their temporal and input component behavior to be identical to the original one-to-one task with one exception: instead of the encoder and decoder temporal components following one another, we find one trajectory is flipped in such a way as to yield an anti-diagonal attention matrix (Fig.~\ref{fig:summary}g).
That is, the last encoder temporal component is aligned with the first decoder temporal component and vice versa.

Second, consider the task of sorting the input alphabetically, e.g. $\{\text{B}, \text{C}, \text{A},  \text{D}\} \to \{\text{A}, \text{B}, \text{C}, \text{D}\}$.
For this example, we expect the network to learn an input-dependent attention matrix that correctly permutes the input sequence.
Since there is no longer a correlation between input and output sequence location, the average attention matrix is very different from that of a random sample, and so we expect the temporal components to insignificantly contribute to the alignment.
Indeed, we find $\dect{s}\cdot\enct{t}$ to be negligible, and instead $\decd{s}\cdot\enci{x}$ dominates the alignment values (Fig.~\ref{fig:summary}h).

\section{Beyond One-to-One Results}

In this section we analyze the dynamics of two tasks that have ``close-to-diagonal'' attention: (1) what we refer to as the extended SCAN  dataset and (2) translation between English and French phrases.
Since we found temporal/input component decomposition to provide little insight into VED dynamics, our focus in this section will be on only the AED and AO architectures.
For both tasks we explore below, parts of the picture we established on the one-to-one task continues to hold.  
However, we will see that in order to succeed at these tasks, both AO and AED must implement additional mechanisms on top of the dynamics we saw for the one-to-one task. 

\begin{figure*}[t]
    \centering
    \includegraphics[width=0.95\textwidth]{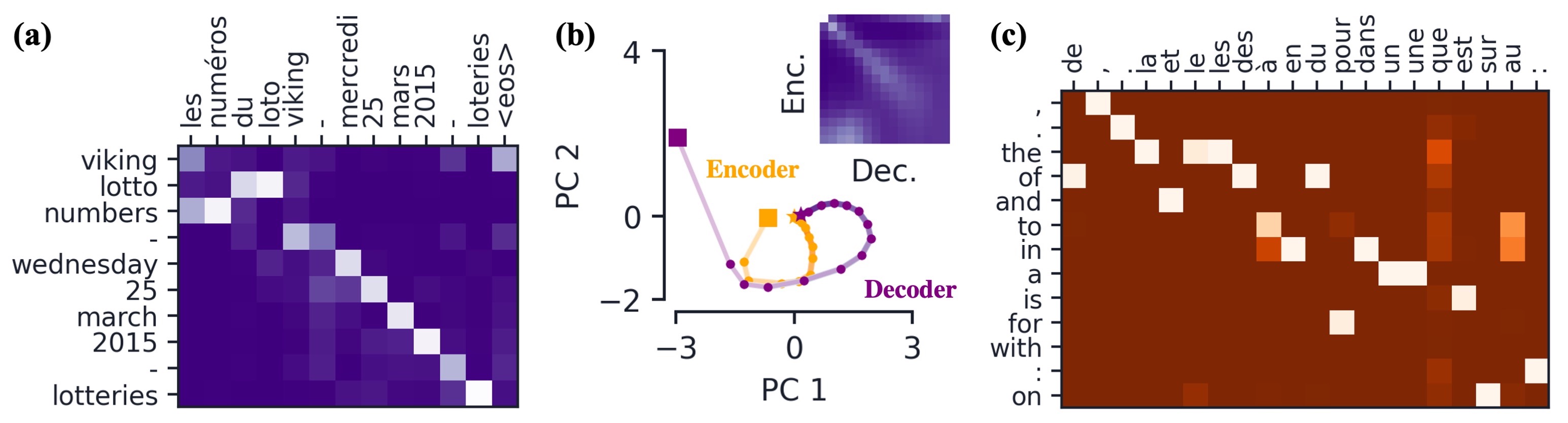}
    \vspace{-1mm}
    \caption{\small
    \textbf{Summary of features for AO trained on English to French translation.} 
    \textbf{(a)} Sample attention matrix. 
    \textbf{(b)} The encoder (orange) and decoder (purple) temporal components, with a square and star marking the first and last time step, respectively. 
    Once again, quantities are projected onto the temporal component PCs.
    The inset shows the attention matrix from the temporal components, i.e. the softmax of $\dect{s} \cdot \enct{t}$.
    \textbf{(c)} The dot product between the most common output word readouts and the most common input word input components, $\enci{x}$.
    }
    \vspace{-4mm}
    \label{fig:translation}
\end{figure*}

\textbf{Extended SCAN (eSCAN)} is a modified version of the SCAN dataset \citep{scan}, in which we randomly concatenate a subset of the phrases to form phrases of length $15$ to $20$ (see \swaptext{SM}{Appendix \ref{app:datasets}} for details).  
The eSCAN tasks is close to one-to-one translation, but is augmented with several additional rules that modify its structure.
For example, a common sequence-to-sequence structure is that a pair of outputs can swap order relative to their corresponding inputs: the English words `green field' translate to `champ vert' in French (with `field' $\leftrightarrow$ `champ'  and   `green' $\leftrightarrow$ `vert'). 
This behavior is present in eSCAN: when the input word `left' follows a verb the output command must first turn the respective direction and then perform said action (e.g. `run left' $\rightarrow$ `LTURN RUN').


The AED and AO models both achieve $\geq98\%$ word accuracy on eSCAN. 
Looking at a sample attention matrix of AED, we see consecutive words in the output phrase tend to attend to the same encoder hidden states at the end of subphrases in the input phrase (Fig.~\ref{fig:scan_qual}a). 
Once again decomposing the AED network's hidden states as in \eqref{eq:hs_temporal_word}, we find the temporal components of the encoder and decoder form curves that mirror one another, leading to an approximately diagonal attention matrix (Fig.~\ref{fig:scan_qual}b).
The delta components are significantly less negligible for this task, as evidence by the fact $\enci{x} + \encd{t}$ aren't nearly as clustered around their corresponding input component (Fig.~\ref{fig:scan_qual}c).
As we will verify later, this is a direct result of the network's use of recurrence, since now hidden states carry information about subphrases, rather than just individual words.

Training the AO architecture on eSCAN, we also observe non-diagonal attention matrices, but in general their qualitative features differ from those of the AED architecture (Fig.~\ref{fig:scan_qual}d). 
Focusing on the subphrase mapping `run twice' $\rightarrow$ `RUN RUN', we see the network learns to attend to the word preceding `twice', since it can no longer rely on recurrence to carry said word's identity forward. 
Once again, the temporal components of the encoder and decoder trace out paths that roughly follow one another (Fig.~\ref{fig:scan_qual}e). 
We see input-delta components cluster around their corresponding input components, indicating the delta components are small (Fig.~\ref{fig:scan_qual}f).
Finally, we again see the readouts of particular outputs align well with the input components of their corresponding input word.

\textbf{English to French Translation} is another example of a nearly-diagonal task. 
We train the AED and AO architectures on this natural language task using a subset of the para\_crawl dataset~\cite{banon-etal-2020-paracrawl} consisting of over 30 million parallel sentences.
To aid interpetation, we tokenize each sentence at the word level and maintain a vocabulary of 30k words in each language; we train on sentences of length up to 15 tokens.  

Since English and French are syntactically similar with roughly consistent word ordering, the attention matrices are in general close to diagonal  (Fig.~\ref{fig:translation}a).
Again, note the presence of features that require off-diagonal attention, such as the flipping of word ordering in the input/output phrases and multiple words in French mapping to a single English word.
Using the decomposition of \eqref{eq:hs_temporal_word}, the temporal components in both AED and AO continue to trace out similar curves (Fig.~\ref{fig:translation}b).
Notably, the alignment resulting from the temporal components is significantly less diagonal, with the diagonal behavior clearest at the beginning of the phrase.
Such behavior makes sense: the presence of off-diagonal structure means, on average, translation pairs become increasingly offset the further one moves into a phrase.
With offsets that increasingly vary from phrase to phrase, the network must rely less on temporal component alignments, which by definition are independent of the inputs.
Finally, we see that the the dot product between the input components and the readout vectors implement the translation dictionary, just as it did for the one-to-one task (Fig.~\ref{fig:translation}c, see below for additional discussion). 

\subsection{A Closer Look at Model Features}
\label{sec:model_features}

\begin{figure*}[t]
    \centering
    \includegraphics[width=0.95\textwidth]{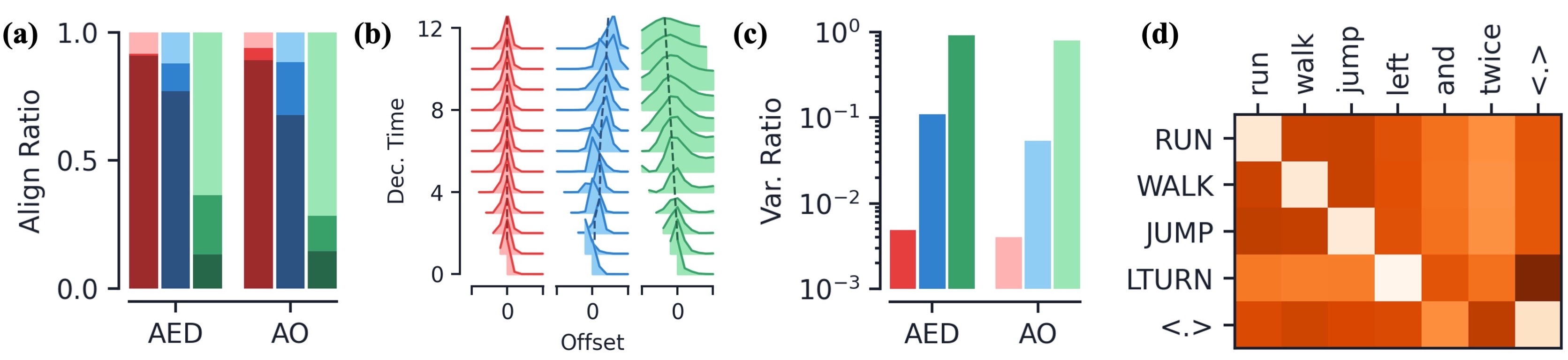}
    \vspace{-1mm}
    \caption{\small
    \textbf{Temporal and input component features.} 
    In the first three plots, the data shown in red, blue, and green corresponds to networks trained on the one-to-one, eSCAN, and English to French translation tasks, respectively.
    \textbf{(a)}  Breakdown of the nine terms that contribute to the largest alignment scores (see \eqref{eq:alignment}) averaged across the entire decoder sequence for each task/architecture combination (see \swaptext{SM}{Appendix~\ref{app:methods}} for details). 
    For each bar, from top to bottom, the alignment contributions from $\dect{s}\cdot\enct{t}$ (dark), $\dect{s}\cdot\enci{x} + \dect{s}\cdot\encd{t}$ (medium), and the remaining six terms (light).
    \textbf{(b)} For the AO architecture, the dot product of the temporal components, $\dect{s}\cdot\enct{t}$, as a function of the offset, $t-s$, shown at different decoder times.
    Each offset is plotted from $[-5, 5]$ and the dotted lines show the theoretical prediction for maximum offset as a function of decoder time, $s$.
    Plots for the AED architecture are qualitatively similar.
    \textbf{(c)} For all hidden states corresponding to an input word, the ratio of variance of $\ench{t}-\enct{t}$ to $\ench{t}$.
    \textbf{(d)} For AO trained on eSCAN, the dot product of input components, $\enci{x}$, with each of the readouts (AED is qualitatively similar).
    }
    \label{fig:scan_quan}
    \vspace{-4mm}
\end{figure*}

As expected, both the AED and AO architectures have more nuanced attention mechanisms when trained on eSCAN and translation.
In this section, we investigate a few of their features in detail.

\textbf{Alignment Approximation.} \hspace{0.5mm}
Recall that for the one-to-one task, we found the alignment scores could be well approximated by $a_{st}\approx \dect{s} \cdot \enct{t}$, which was agnostic to the details of the input sequence.
For eSCAN, the $\dect{s} \cdot \enct{t}$ term is still largely dominant, capturing $>77\%$ of $a_{st}$ in the AED and AO networks (Fig.~\ref{fig:scan_quan}a).
A better approximation for the alignment scores is $a_{st} \approx \dect{s} \cdot \enct{t} + \dect{s} \cdot \enci{x} + \dect{s} \cdot \encd{t}$, i.e. we include two additional terms on top of what was used for one-to-one.
Since $\enci{x}$ and $\encd{t}$ are dependent upon the input sequence, this means the alignment has non-trivial input dependence, as we would expect.
In both architectures, we find this approximation captures $>87\%$ of the top alignment scores. 
For translation, we see the term $\dect{s} \cdot \enct{t}$ makes up a significantly smaller portion of the alignment scores, and in general we find none of the nine terms in \eqref{eq:alignment} dominate above the rest (Fig.~\ref{fig:scan_quan}a).
However, at early times in the AED architecture, we again see $\dect{s} \cdot \enct{t}$ is the largest contribution to the alignment.
As mentioned above, this matches our intuition that words at the start of the encoder/decoder phrase have a smaller offset from one another than later in the phrase, so the network can rely more on temporal components to determine attention.


\textbf{Temporal Component Offset.} \hspace{1mm}
For the one-to-one task, the input sequence length was always equal to the output sequence length, so the temporal components were always peaked at $s=t$ (Fig.~\ref{fig:scan_quan}b).
In eSCAN, the input word `and' has no corresponding output, which has a non-trivial effect on how the network attends since its appearance means later words in input phrase are offset from their corresponding output word. 
This effect also compounds with multiple occurrences of `and' in the input.
The AED and AO networks learn to handle such behavior by biasing the temporal component dot product, $\dect{s}\cdot\enct{t}$, the dominant alignment contribution, to be larger for time steps $t$ further along in the encoder phrase, i.e. $t>s$ (Fig.~\ref{fig:scan_quan}b).
It is possible to compute the average offset of input and output words in eSCAN training set, and we see the maximum of  $\dect{s}\cdot\enct{t}$ follow this estimate quite well.
Similarly, in our set of English to French translation phrases, we find the French phrases to be on average $\sim 20\%$ longer than their English counterparts.
This results in the maximum of $\dect{s}\cdot\enct{t}$ to gradually move toward $t<s$, e.g. on average the decoder attends to earlier times in the encoder (Fig.~\ref{fig:scan_quan}b). 
Additionally, note the temporal dot product falls off significantly slower as a function of offset for later time steps, indicating the drop off for non-diagonal alignments is smaller and thus it is easier for the network to off-diagonally attend.

\textbf{Word Variance.} \hspace{1mm}
The encoder hidden states in the one-to-one task had a negligible delta component, so the hidden states could be approximated as $\ench{t}\approx\enct{t}+\enci{x}$. 
By definition, $\enci{x}$ is constant for a given input word, so the variance in the hidden states corresponding to a given input word is primarily contained in the temporal component (Fig.~\ref{fig:scan_quan}c).
Since the temporal component is input-independent, this led to a clear understanding of how all of a network's hidden states evolve with time and input.
In the AED and AO architectures trained on eSCAN, we find the variance of the input word's hidden states drops by $90\%$ and $95\%$ when the temporal component is subtracted out, respectively (Fig.~\ref{fig:scan_quan}c).
Meanwhile, in translation, we find the variance only drops by $8\%$ and $25\%$ for the AED and AO architectures, indicating there is significant variance in the hidden states beyond the average temporal evolution and thus more intricate dynamics.
 
\textbf{Input/Readout Alignment.} \hspace{1mm}
Lastly, recall we saw that in the one-to-one case the input components' alignment with readouts implemented the translation dictionary (Figs.~\ref{fig:summary}b, d).
For eSCAN, the dot product of a given readout is again largest with the input component of its corresponding input word, e.g. the readout corresponding to `RUN' is maximal for the input component of `run' (Fig.~\ref{fig:scan_quan}d).
Notably, words that produce no corresponding output such as `and' and `twice' are not the maximal in alignment with any readout vector. 
Similarly, for translation, we see the French-word readouts have the largest dot product their translated English words (Fig.~\ref{fig:translation}c).
For example, the readouts for the words `la', `le', and `les', which are the gendered French equivalents of `the',  all have maximal alignments with $\enci{\text{the}}$.

\subsection{A Closer Look at Dynamics}
\label{sec:eSCAN_detail}

\begin{figure*}[t]
    \centering
    \includegraphics[width=0.95\textwidth]{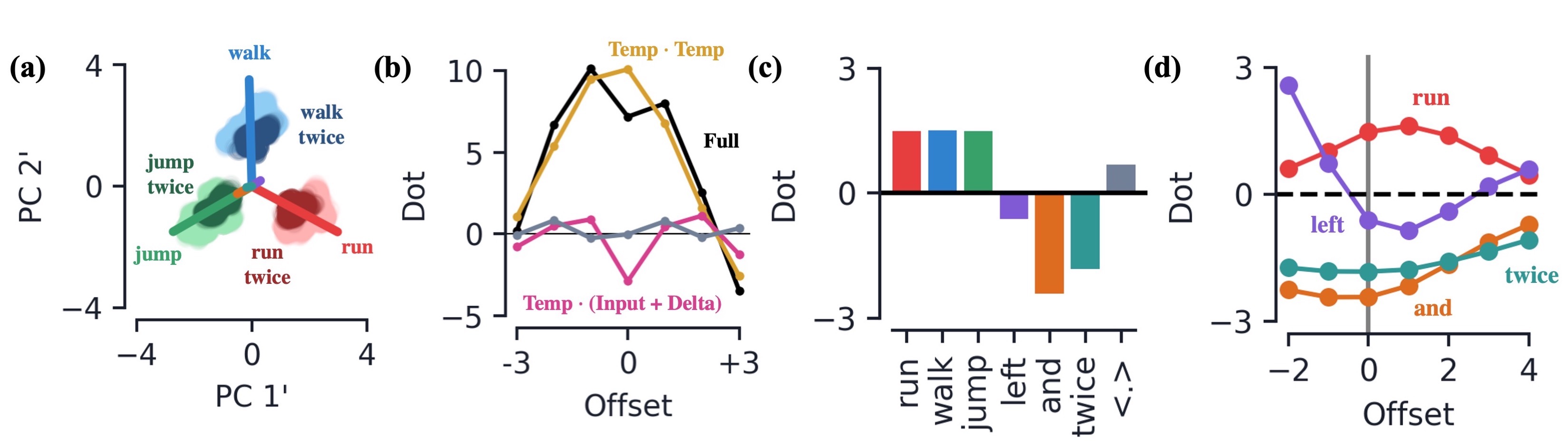}
    \vspace{-3mm}
    \caption{\small
    \textbf{How AO and AED networks implement off-diagonal attention in the eSCAN dataset.} 
    \textbf{(a)} For AED, the input-delta components for various words and subphrases.
    \textbf{(b)} For AO, the alignment values, $a_{st}$, are shown in black when the input word `twice' is at $t=s$.
    Three contributions to the alignment, $\dect{s}\cdot\enct{t}$ (gold), $\dect{s}\cdot\enci{x} + \dect{s}\cdot\encd{t}$ (pink), and  $a_{st} - \dect{s}\cdot\ench{t}$ (grey) are also plotted.
    To keep the offset between `twice' and the output location of the repeated word constant, this plot was generated on a subset of eSCAN with $T=S$, but we observe the same qualitative features when $T \geq S$.
    \textbf{(c)} The dot product between $\enci{x}+\encd{t}$ and the decoder's temporal component, $\dect{s}$, for $t=s$.
    \textbf{(d)} How the dot product of $\enci{x}+\encd{t}$ and $\dect{s}$ changes as a function of their offset, $t-s$, for a few select input words. 
    The vertical gray slice represents the data in (c) and the input word colors are the same.
    }
    \label{fig:closer_scan}
    \vspace{-4mm}
\end{figure*}

In this section, we leverage the temporal and input component decomposition to take a closer look at how networks trained on the eSCAN dataset implement particular off-diagonal attentions.
Many of the sequence translation structures in eSCAN are seen in realistic datasets, so we this analysis will give clues toward understanding the behavior of more complicated sequence-to-sequence tasks.

A common structure in sequence-to-sequence tasks is when an output word is modified by the words preceding it. 
For example, the phrases `we run` and `they run' translate to `nous courrons' and `ils courent' in French, respectively (with the second word in each the translation of `run'). 
We can study this phenomenon in eSCAN since the word `twice' tells the network to repeat the command just issued two times, e.g. `run twice' outputs to `RUN RUN'. 
Hence, the output corresponding to the input `twice' changes based on other words in the phrase.

Since an AED network has recurrence, when it sees the word `twice' it can know what verb preceded it.
Plotting input-delta components, we see the RNN outputs `twice' hidden states in three separate clusters separated by the preceding word (Fig.~\ref{fig:closer_scan}a). 
Thus for an occurrence of `twice' at time step $t$, we have $\enci{\text{twice}} + \encd{t} \approx \enci{\text{verb}} + \encd{t-1}$.
For example, this means the AED learns to read in `run twice' approximately the same as `run run'.
This is an example of the network learning context.

AO has no recurrence, so it can't know which word was output before `twice'. 
Hence, unlike the AED case, all occurrences of `twice' are the same input-delta component cluster regardless of what word preceded it. 
Instead, it has to rely on attending to the word that modifies the output, which in this case is simply the preceding word (Fig.~\ref{fig:scan_qual}d). 
As mentioned in Sec.~\ref{sec:model_features}, for the eSCAN task we find the alignment to be well approximated by $a_{st} \approx \dect{s} \cdot \ench{t}$. 
When the word 'twice' appears in the input phrase, we find $\dect{s}\cdot \enci{\text{twice}} + \dect{s}\cdot \encd{t} <0$ for $s=t$ (Fig.~\ref{fig:closer_scan}b).
This decreases the value of the alignment $a_{s,s}$, and so the decoder instead attends to the time step with the second largest value of $\dect{s} \cdot \enct{t}$, which the network has learned to be $t=s-1$.
Hence, $a_{s, s-1}$ is the largest alignment, corresponding to the time step before `twice' with the verb the network needs to output again. 
Unlike the one-to-one case, the encoder input-delta and the decoder temporal components are no longer approximately orthogonal to one another (Fig.~\ref{fig:closer_scan}c). 
In the case of `twice', $\enci{\text{twice}} + \encd{t}$ is partially antialigned with the temporal component, yielding a negative dot product.

This mechanism generalizes beyond the word 'twice': in eSCAN we see input-delta components of several input words are no longer orthogonal to the decoder's temporal component (Fig.~\ref{fig:closer_scan}c).
Like `twice', the dot product of the input-delta component for a given word with its corresponding temporal component determines how much its alignment score is increased/decreased.
For example, we see $\enci{\text{and}} + \encd{t}$ has a negative dot product with the temporal component, meaning it leans away from its corresponding temporal component.
Again, this make sense from eSCAN task: the word `and' has no corresponding output, hence it never wants to be attended to by the decoder.

Perhaps contradictory to expectation, $\enci{\text{left}} + \encd{t}$ has a negative dot product with the temporal component.
However, note that the alignment of $\enci{x} + \encd{t}$ with the $\dech{s}$ is dependent on both $t$ and $s$. 
We plot the dot products of $\enci{x} + \encd{t}$ and $\dech{s}$ as a function of their offset, defined to be the $t-s$  (Fig.~\ref{fig:closer_scan}d).
Notably, $\enci{\text{left}} + \encd{t}$ has a larger dot product for larger offsets, meaning it increases its alignment when $t > s$. 
This makes sense from the point of view that the word `left' is always further along in the input phrase than its corresponding output `LTURN', and this offset is only compounded by the presence of the word `and'. 
Thus, the word `left' only wants to get noticed if it is ahead of the corresponding decoder time step, otherwise it hides.
Additionally, the words `and` and `twice` have large negative dot products for all offsets, since they never want to be the subject of attention.

\section{Discussion}

In this work, we studied the hidden state dynamics of sequence-to-sequence tasks in architectures with recurrence and attention.
We proposed a decomposition of the hidden states into parts that are input- and time-independent and showed when such a decomposition aids in understanding the behavior of encoder-decoder networks.

Although we have started by analyzing translation tasks, it would be interesting to understand how said decomposition works on different sequence-to-sequence tasks, such as speech-to-text.
Additionally, with our focus on the simplest encoder-decoder architectures, it is important to investigate how much the observed dynamics generalize to more complicated network setups, such as networks with bidirectional RNNs or multiheaded and self-attention mechanisms.
Our analysis of the attention-only architecture, which bears  resemblance to the transformer architecture, suggests that a similar dynamical behavior may also hold for the Transformer, hinting at the working mechanisms behind this popular non-recurrent architecture.

\begin{ack}

We thank Ankush Garg for collaboration during the early part of this work.
None of the authors receive third-party funding/support during the 36 months prior to this submission or had competing interests. 

\end{ack}

\bibliography{seq2seq}
\bibliographystyle{icml2021}


\appendix

\section{Additional Details}

\begin{figure*}[ht]
    \centering
    \includegraphics[width=0.95\textwidth]{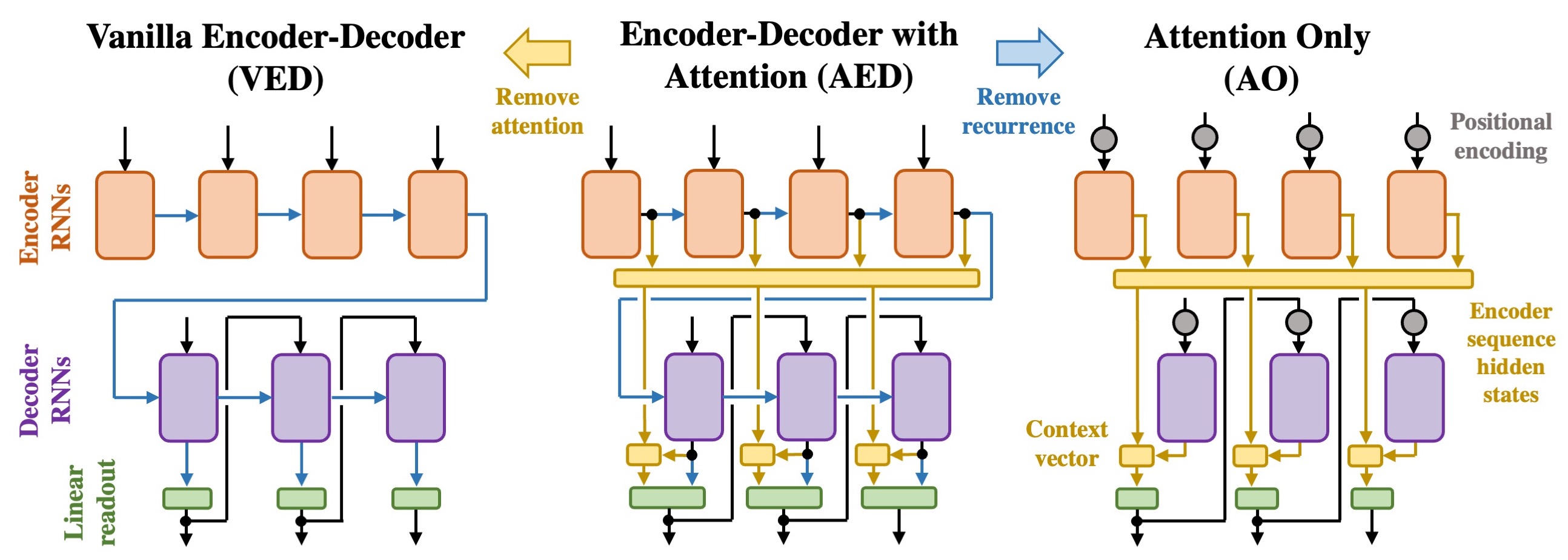}
    \vspace{-3mm}
    \caption{\small
    \textbf{Comparison of the three primary architectures used in this work and the relation between them.} 
    The three architectures are vanilla encoder-decoder (VED), encoder-decoder with attention (AED), and attention only (AO).  The encoder RNNs, decoder RNNs, and linear readout layer are showing in orange, purple, and green, respectively.
    Recurrent connections between RNNs are shown in blue, attention-based connections and computational blocks are shown in gold.
    For AO, the grey circles represent locations where positional encoding is added to the inputs.
    Note AED's linear readout layer takes in both the context vector from attention as well as the decoder's output.
    }
    \label{fig:arch}
\end{figure*}

In this section we provide additional details regarding the architectures, temporal-input component decomposition, datasets, RNNs, and training used in this work.

As a reminder, the encoder and decoder hidden states are denoted by $\ench{t}$ and $\dech{s}$, respectively, while inputs to the encoder and decoder are denoted by $\mathbf{x}_t^\text{E}$ and $\mathbf{x}_s^\text{D}$.

\subsection{Architectures}
\label{app:archs}

A summary of the three architectures we focus on---vanilla encoder-decoder, encoder-decoder with attention, and attention only---in this work is shown in Fig.~\ref{fig:arch}.
Intuitively, the architectures are related as follows: AED has attention and recurrence, VED and AO are the same as AED with the attention and recurrence removed, respectively (for AO, we also add positional encoding).

\subsubsection*{Vanilla Encoder Decoder}
The encoder and decoder update expression are 
\begin{align}
    \ench{t} = F_\text{E}(\ench{t-1}, \mathbf{x}_t^\text{E})\,, \qquad \dech{s} = F_\text{D}(\dech{s-1},\mathbf{x}_s^\text{D})\, ,
\end{align}
respectively. 
Here, $F_\text{D}$ and $F_\text{E}$ are the functions that implement the hidden state updates, which in this work are each one of three modern RNN architectures: LSTMs~\citep{HochSchm97}, GRUs~\citep{GRU}, or UGRNNs~\citep{UGRNN}.
The final encoder hidden state is the decoder's initial hidden state, so that $\dech{0} = \ench{T}$. 
The decoder hidden states are passed through a linear layer to get the output logits at each time step, $\mathbf{y}_s = \mathbf{W}\dech{s} + \mathbf{b}$, with the following decoder input, $\mathbf{x}_{s+1}^\text{D}$, determined by the word corresponding to the maximum output logit, $\text{argmax}(\mathbf{y}_s)$.

\subsubsection*{Encoder Decoder with Attention}

The encoder-decoder with attention architecture is identical to the VED architecture above with a simple attention mechanism added \citep{bahdanau2014neural, luong2015effective}.
For time step $s$ of the decoder, we compute a  context vector $\mathbf{c}_s$, a weighted sum of encoder hidden states,
\begin{align}
\label{eq:context_vector}
    \mathbf{c}_s := \sum_{t=1}^T \alpha_{st} \ench{t}\, ,\qquad \alpha_{st} :=\frac{e^{a_{st}}}{\sum_{t^{\prime}=1}^T e^{a_{st^\prime}}}\,.
\end{align}
Here, $\bm{\alpha}_{t} := \text{softmax}\left(a_{1t}, \ldots, a_{St} \right)$ is the $t$-th column of the \emph{attention matrix} and $a_{st}:=\dech{s}\cdot \ench{t}$ the \emph{alignment} between a given decoder and encoder hidden state. 
Furthermore, the outputs are now determined by passing \emph{both} the decoder hidden state and the context vector through the linear layer, i.e. $\mathbf{y}_s = \mathbf{W}[\dech{s}, \mathbf{c}_s] + \mathbf{b}$, where $[\cdot,\cdot]$ denotes concatenation \cite{luong2015effective}. 

\subsubsection*{Attention Only}

Attention only is identical to the AED network above, but simply eliminates the recurrent information passed from one RNN cell to the next.
Since this eliminates any sense of temporal ordering in the sequences, we also add fixed positional encoding vectors \citep{vaswani2017attention}, $\mathbf{p}_t^\text{E}$ and $\mathbf{p}_s^\text{D}$ to the encoder and decoder inputs.
Together, this means the hidden state update expressions are now 
\begin{align}
    \ench{t} = F_\text{E}(\mathbf{0},\mathbf{x}_t^\text{E} + \mathbf{p}_t^\text{E})\, , \qquad \dech{s} = F_\text{D}(\mathbf{0},\mathbf{x}_s^\text{D} + \mathbf{p}_s^\text{D}) \,. 
\end{align}
Note the RNN functions $F_\text{E}$ and $F_\text{D}$ simply act as feedforward networks in this setting.
Lastly, the output logits are now determined solely from the context vector, $\mathbf{y}_s = \mathbf{W}\mathbf{c}_s + \mathbf{b}$.

Although using gated RNNs cells as feedforward networks is fairly non-standard, our primary motivation is to keep the AED and AO architectures as similar as possible in order to isolate the differences that arise from recurrence and positional encoding.
Below we discuss a non-gated feedforward variant that we also briefly investigate.

Note the elimination of recurrence means the entire encoder hidden states sequence can be computed in parallel.  
This architecture is meant to be a simplified model of a Transformer \citep{vaswani2017attention}. 
The hidden states output by the RNN simultaneous fill the role of the usual keys, queries, and values of a Transformer.
Additionally, there is no self-attention mechanism, only a single ``head'' of attention, and no residual connections.

\paragraph{Positional Encoding}
For a $d$-dimensional embedding dimension at time $t$, we add the vector $\mathbf{p}_t^\text{E}$ with $i=0,\ldots, d-1$ components:
\begin{align}
p_{t,i}^\text{E} =\begin{cases}
\sin\left(\frac{t}{\tau^{i/d}}\right) & i\,\text{even}\\
\cos\left(\frac{t}{\tau^{(i-1)/d}}\right) & i\,\text{odd}
\end{cases}    
\end{align}
with $\tau$ some temporal scale that should be related to the phrase length. 
This is the same positional encoding used in \citet{vaswani2017attention}, and we use the same encoding for both the encoder ($\mathbf{p}_t^\text{E}$) and decoder ( $\mathbf{p}_t^\text{D}$).

\paragraph{Non-Gated Variant}

As mentioned above, we use a gated-RNN with its recurrence cut as a feedforward network.
Since this is fairly non-standard, we also verify some of our results on non-gated feedforward architectures.
The non-gated variant of AO uses a the following hidden-state updates
\begin{subequations}
\begin{align}
    \ench{t} =  F^\prime_\text{E}(\mathbf{x}^\text{E}_t) & := \text{tanh}\left(\mathbf{W}^\text{E} \mathbf{x}^\text{E}_t + \mathbf{b}^\text{E}\right)\, , \\
    \dech{s} = F^\prime_\text{D}(\mathbf{x}^\text{D}_s) & := \text{tanh}\left(\mathbf{W}^\text{D} \mathbf{x}^\text{D}_s + \mathbf{b}^\text{D}\right)\, ,
\end{align}
\end{subequations}
with tanh acting pointwise. 
This architecture is identical to the version of AO used above, but the hidden state updates are now
\begin{align}
    \ench{t} = F^\prime_\text{E}(\mathbf{x}_t^\text{E} + \mathbf{p}_t^\text{E})\, , \qquad \dech{s} = F^\prime_\text{D}(\mathbf{x}_s^\text{D} + \mathbf{p}_s^\text{D}) \,. 
\end{align}
Below, we show that we find the qualitative results of this network are the same as the gated version, and thus our results do not seem to be dependent upon the gating mechanisms present in the feedforward networks of the AO architecture.

\subsection{Temporal and Input Components}
\label{app:methods}

We define the \emph{temporal components} to be the average hidden state at a given time step.
In practice, we estimate such averages using a test set of size $M$, so that the temporal components are given by
\begin{subequations}
\begin{align}
\enct{t} \approx & \frac{\sum_{
\alpha=1}^M \mathbf{1}_{\leq\text{EoS}, \alpha} \ench{t, \alpha}}{\sum_{
\beta=1}^M \mathbf{1}_{\leq\text{EoS}, \beta}}\, , \\
\dect{s} \approx & \frac{\sum_{\alpha=1}^M \mathbf{1}_{\leq\text{EoS}, \alpha} \dech{s, \alpha}}{\sum_{
\beta=1}^M \mathbf{1}_{\leq\text{EoS}, \beta}}\, ,
\end{align}
\end{subequations}
with $\ench{t, \alpha} $ the encoder hidden state of the $\alpha$th sample and $\mathbf{1}_{\leq\text{EoS}, \alpha}$ is a mask that is zero if the $\alpha$th sample is beyond the end of sentence.
Next, we define the encoder \emph{input components} to be the average of $\ench{t}-\enct{t}$ for all hidden states that immediately follow a given input word (and similarly for the decoder input components).
Once again, we estimate the input components using a test set of size $M$,
\begin{subequations}
\begin{align}
    \bm{\chi}^\text{E}\left(\mathbf{x}_{t, \alpha}\right) \approx & \frac{\sum_{\beta=1}^M \sum_{t^\prime=1}^T \mathbf{1}_{\mathbf{x}_{t,\alpha}, \mathbf{x}_{t^\prime,\beta}} \left(\ench{t^\prime,\beta} - \enct{t^\prime}\right)}{\sum_{\gamma=1}^M \sum_{t^{\prime\prime}=1}^T \mathbf{1}_{\mathbf{x}_{t,\alpha}, \mathbf{x}_{t^{\prime\prime},\gamma}}}\,, \\
    \bm{\chi}^\text{D}\left(\mathbf{x}_{s, \alpha}\right) \approx & \frac{\sum_{\beta=1}^M \sum_{s^\prime=1}^S \mathbf{1}_{\mathbf{x}_{s,\alpha}, \mathbf{x}_{s^\prime,\beta}} \left(\dech{s^\prime,\beta} - \dect{s^\prime}\right)}{\sum_{\gamma=1}^M \sum_{s^{\prime\prime}=1}^S \mathbf{1}_{\mathbf{x}_{s,\alpha}, \mathbf{x}_{s^{\prime\prime},\gamma}}}\,,
\end{align}
\end{subequations}
where $\mathbf{1}_{\mathbf{x}_{t,\alpha}, \mathbf{x}_{t^\prime,\beta}}$ is a mask that is zero if $\mathbf{x}_{t,\alpha} \ne \mathbf{x}_{t^\prime,\beta}$ and we have temporally suppressed the superscripts on the inputs for brevity.
With the above definitions, we can decompose encoder and decoder hidden states resulting from the $\alpha$th sample as
\begin{subequations}
\begin{align}
    \ench{t,\alpha} & = \enct{t} + \bm{\chi}^\text{E}\left(\mathbf{x}^\text{E}_{t, \alpha}\right) + \encd{t,\alpha}\,, \\
    \dech{s,\alpha} & = \dect{s} + \bm{\chi}^\text{D}\left(\mathbf{x}^\text{D}_{s, \alpha}\right) + \decd{s,\alpha}\,,
\end{align}
\end{subequations}
with the \emph{delta components} defined to be whatever is leftover in the hidden state after subtracting out the temporal and input components,
\begin{subequations}
\begin{align}
    \encd{t,\alpha} & := \ench{t,\alpha} - \enct{t} - \bm{\chi}^\text{E}\left(\mathbf{x}^\text{E}_{t, \alpha}\right) \,, \\
    \decd{s,\alpha} & := \dech{s,\alpha} - \dect{s} - \bm{\chi}^\text{D}\left(\mathbf{x}^\text{D}_{s, \alpha}\right) \,.
\end{align}
\end{subequations}
In the main text we use the shorthand $\enci{x} = \bm{\chi}^\text{E}\left(\mathbf{x}^\text{E}_{t, \alpha}\right)$ and $\deci{y} = \bm{\chi}^\text{D}\left(\mathbf{x}^\text{D}_{s, \alpha}\right)$ (since for the decoder, the previous time step's output, $\textbf{y}_{s-1}$ is passed as the next input).
We will often suppress the batch index, so altogether the decomposition is written in the main text as
\begin{subequations}
\begin{align}
    \ench{t} & = \enct{t} + \enci{x} + \encd{t}\,, \\
    \dech{s} & = \dect{s} + \deci{y} + \decd{s}\,.
\end{align}
\end{subequations}

The intuition behind this decomposition is that it is an attempt to isolate the temporal and input behavior of the network's hidden state updates $F_\text{E}$ and $F_\text{D}$.
This partially motivated by the fact that, \emph{if} $F_\text{E}$ were linear, then the encoder hidden state update for AO could be written as
\begin{align}
    \ench{t} = F_\text{E}(\mathbf{0},\mathbf{x}_t^\text{E} + \mathbf{p}_t^\text{E}) = F_\text{E}(\mathbf{0},\mathbf{x}_t^\text{E}) + F_\text{E}(\mathbf{0},\mathbf{p}_t^\text{E})\,.
\end{align}
Notably, the first term is only dependent upon the input and the second term is only dependent upon the sequence index (through the positional encoding).
In this case, the temporal and input component would \emph{exactly} capture the time and input dependence of the hidden states, respectively.
Of course, $F_\text{E}$ is not in general linear, but we still find such a decomposition useful for interpretation.

\paragraph{Alignment}

Using the above decomposition, we can write the alignment as a sum of nine terms:
\begin{subequations}
\begin{align}
    a_{st} & = \left(\dect{s} + \deci{y} + \Delta\dech{s}\right) \cdot \left(\enct{t} + \enci{x} + \Delta\ench{t}\right) \nonumber \\
    & = \dect{s} \cdot \enct{t} + \dect{s} \cdot\enci{x} + \dect{s} \cdot\encd{t} \nonumber \\ 
    & \quad + \deci{y} \cdot \enct{t} + \deci{y} \cdot\enci{x} + \deci{y} \cdot\encd{t} \nonumber \\
    & \quad + \decd{s} \cdot \enct{t} + \decd{s} \cdot\enci{x} + \decd{s} \cdot\encd{t} \\
    & := \sum_{I=1}^9 a_{st}^{(I)}\,,
\end{align}
\end{subequations}
with $a_{st}^{(I)}$ for $I=1,\ldots,9$ defined as the nine terms which sequentially appear after the second equality (i.e. $a_{st}^{(1)}:= \dect{s} \cdot \enct{t}$, $a_{st}^{(2)}:= \dect{s} \cdot \enci{x}$, and so on).

In the main text, we measure the breakdown of the alignment scores from each of the nine terms. 
Define the contributions from one of the nine terms as
\begin{align}
    A_{st}^{(I)} := \frac{\left|a_{st}^{(I)} \right|}{\sum_{J=1}^9 \left| a_{st}^{(J)}\right|} \, ,
\end{align}
where the absolute values are necessary because contributions to the dot product alignment can be positive or negative.


\subsection{Datasets}
\label{app:datasets}

\paragraph{One-to-One Dataset}
This is a simple sequence to sequence task consisting variable length phrases with input and output words that are in one-to-one correspondence. 
At each time step, a word from a word bank of size $N$ is randomly chosen (uniformly), and as such there is no correlation between words at different time steps.
The length of a given input phrase is predetermined and also drawn a uniform distribution.
After an input phrase is generated, the corresponding output phrase is created by individually translating each word.
All input words translate to a unique output word and translations are solely dependent upon the input word. 
An example one-to-one dataset would be converting a sequence of letters to their corresponding position in the alphabet, $\{\text{B}, \text{A}, \text{C}, \text{A}, \text{A}\} \to \{2, 1, 3, 1, 1\}$.
Note the task of simply repeating the input phrase as the output is also one-to-one.

Due to the small vocabulary size, one-hot encoding is used for input phrases.
Train and test sets are generated dynamically. 

\paragraph{Extended SCAN}

\begin{table}[t]
    \centering
    \begin{tabular}{lll}
    \toprule
    \textbf{Input Phrase} & \textbf{Output Phrase} \\
    \midrule
    run $\langle$.$\rangle$ jump $\langle$.$\rangle$ walk $\langle$.$\rangle$   &  RUN $\langle$.$\rangle$ JUMP $\langle$.$\rangle$ WALK $\langle$.$\rangle$ \\
    run left $\langle$.$\rangle$   &  LTURN RUN $\langle$.$\rangle$\\
    run twice $\langle$.$\rangle$ jump $\langle$.$\rangle$  &  RUN RUN $\langle$.$\rangle$ JUMP $\langle$.$\rangle$ \\
    run and jump $\langle$.$\rangle$  &  RUN JUMP $\langle$.$\rangle$\\
    \bottomrule
    \end{tabular}
    \caption{Extended SCAN example phrases.}
    \label{tab:eSCAN}
\end{table}

Extended SCAN (eSCAN) is a modified version of the SCAN dataset \cite{scan}. 
The SCAN dataset is a sequence-to-sequence task consisting of translating simple input commands into output actions.
SCAN consists of roughly 20,000 phrases, with a maximum input and output phrase lengths of 9 and 49, respectively. 
A few relevant example phrases of eSCAN are shown in Table \ref{tab:eSCAN}.

The eSCAN dataset modifies SCAN in two ways:
\begin{enumerate}
    \item It takes only a subset of SCAN in which input phrases solely consist of a chosen subset of SCAN words.
    This allows us to isolate particular behaviors present in SCAN as well as eliminate certain word combinations that would require far-from diagonal attention mechanisms (e.g. it allows us to avoid the input subphrase `run around thrice' that yields an output subphrase of length $24$). 
    \item After a subset of SCAN phrases has been chosen, phrases are randomly drawn (uniformly) and concatenated together until a phrase of the desired length range is created.
    Individual SCAN phrases are separated by a special word token (a period).
    This allows us to generate phrases that are much longer than the phrases encountered in the SCAN dataset and also control the variance of phrase lengths.
\end{enumerate}
The eSCAN dataset allows us to gradually increase a phrase's complexity while having control over phrase lengths.
At its simplest, eSCAN can also be one-to-one if one restricts to input phrases with some combination of the words `run', `walk', `jump', and `look'.

Throughout the main text, we use the subset of SCAN that contains the words `run', `walk', `jump', `and', `left', and `twice' with lengths ranging from 10 to 15.
Furthermore, we omit combinations that require a composition of rules, e.g. the phrase `jump left twice' that requires the network to both understand the reversing of `jump left' and the repetition of `twice'. 
Although it would be interesting to study if and how the network learns such compositions, we leave such studies for future work.
This results in 90 distinct subphrases prior to concatenation, with a maximum input and output length of 5 and 4, respectively.
Post concatenation, there are over a million distinct phrases for phrases of length 10 to 15.
Once again, one-hot encoding is used for the input and output phrases and train/test sets are generated dynamically.

\paragraph{Translation Dataset}
Our natural-language translation dataset is the ParaCrawl Corpus, or Web Scale Parallel Corpora for European Languages~\citep{banon-etal-2020-paracrawl}; we train models to translate between English and French, using the release of the dataset available in TensorFlow Datasets.  This release of ParaCrawl features 31,374,161 parallel English/French sentences; as we train our models for 30,000 steps using a batch size of 64, we do not encounter all examples during training.

To aid interpretability, we tokenize the dataset at the word level, by first converting all characters to lowercase, separating punctuation from words, and splitting on whitespace.  Using 10 million randomly chosen sentences from the dataset, we build a vocabulary consisting of the 30,000 most commonly occurring words.  We filter sentences which are longer than 15 tokens.

\subsection{Recurrent Neural Networks}
The three types of RNNs we use in this work are specified below. 
$\mathbf{W}$ and $\mathbf{b}$ represent trainable weight matrices and bias parameters, respectively, and $\mathbf{h}_t$ denotes the hidden state at timestep $t$ (representing either the encoder or decoder).
All other vectors ($\mathbf{c}_t, \mathbf{g}_t, \mathbf{r}_t, \mathbf{i}_t, \mathbf{f}_t$) represent intermediate quantities at time step $t$; $\sigma(\cdot)$ represents a pointwise sigmoid nonlinearity; and $f(\cdot)$ is the pointwise $\text{tanh}$ nonlinearity. 

\paragraph{Gated Recurrent Unit (GRU)}
The hidden state update expression for the GRU \cite{GRU} is given by
\begin{subequations}
\begin{align}
\mathbf{h}_t = \mathbf{g}_t \cdot \mathbf{h}_{t-1} + (1 - \mathbf{g}_t) \cdot \mathbf{c}_t\, ,
\end{align}
with
\begin{align}
\mathbf{c}_t &= f\left(\mathbf{W}^\text{ch} (\mathbf{r} \cdot \mathbf{h}_{t-1}) + \mathbf{W}^\text{cx} \mathbf{x}_t + \mathbf{b}^\text{c}\right) \, ,\\
\mathbf{g}_t &= \sigma\left(\mathbf{W}^\text{gh} \mathbf{h}_{t-1} + \mathbf{W}^\text{gx} \mathbf{x}_t + \mathbf{b}^\text{g}\right) \, ,\\
\mathbf{r}_t &= \sigma\left(\mathbf{W}^\text{rh} \mathbf{h}_{t-1} + \mathbf{W}^\text{rx} \mathbf{x}_t + \mathbf{b}^\text{r}\right)\, ,
\end{align}
\end{subequations}

\paragraph{Update-Gate RNN (UGRNN)}
The hidden state update expression for the UGRNN \cite{UGRNN} is given by 
\begin{subequations}
\begin{align}
\mathbf{h}_t = \mathbf{g}_t \cdot \mathbf{h}_{t-1} + (1 - \mathbf{g}_t) \cdot \mathbf{c}_t\, ,
\end{align}
with
\begin{align}
\mathbf{c}_t &= f\left(\mathbf{W}^\text{ch} \mathbf{h}_{t-1} + \mathbf{W}^\text{cx} \mathbf{x}_t + \mathbf{b}^\text{c} \right) \, , \\
\mathbf{g}_t &= \sigma\left(\mathbf{W}^\text{gh} \mathbf{h}_{t-1} + \mathbf{W}^\text{gx} \mathbf{x}_t + \mathbf{b}^\text{g}\right) \, ,
\end{align}
\end{subequations}

\paragraph{Long-Short-Term-Memory (LSTM)}
Unlike the GRU and the UGRNN, the LSTM \cite{HochSchm97} transfers both a ``hidden state'' and a cell state from one time step to the next.
In order to cast the LSTM update expressions into the same form as the GRU and UGRNN, we define its hidden state to be 
\begin{subequations}
\begin{align}
\mathbf{h}_t = \left[\mathbf{c}_t, \tilde{\mathbf{h}}_t\right] \, ,
\end{align}
with the update expression given by
\begin{align}
\tilde{\mathbf{h}}_t &= f(\mathbf{c}_t) \cdot \sigma\left(\mathbf{W}^\text{hh} \mathbf{h}_{t-1} + \mathbf{W}^\text{hx} \mathbf{x}_{t} + \mathbf{b}^\text{h}\right) \, ,\\
\mathbf{c}_t &= \mathbf{f}_t \cdot \mathbf{c}_{t-1} + \mathbf{i} \cdot \sigma\left(\mathbf{W}^\text{ch} \tilde{\mathbf{h}}_{t-1} + \mathbf{W}^\text{cx} \mathbf{x}_t + \mathbf{b}^\text{c}\right) \, , \\
\mathbf{i}_t &= \sigma\left(\mathbf{W}^\text{ih} \mathbf{h}_{t-1} + \mathbf{W}^\text{ix} \mathbf{x}_t + \mathbf{b}^\text{i}\right) \, ,\\
\mathbf{f}_t &= \sigma\left(\mathbf{W}^\text{fh} \mathbf{h}_{t-1} + \mathbf{W}^\text{fx} \mathbf{x}_t + \mathbf{b}^\text{f}\right) \, .
\end{align}
\end{subequations}
For the LSTM, we only use $\tilde{\mathbf{h}}^\text{E}_t$ and $\tilde{\mathbf{h}}^\text{D}_s$ for the determination of the context vector and as the decoder output that is passed to the readout. That is, for AED, $\mathbf{y}_s = \mathbf{W}[\tilde{\mathbf{h}}^\text{D}_s, \mathbf{c}_s] + \mathbf{b}$ with $\mathbf{c}_s$ the same as \eqref{eq:context_vector} with all $\ench{t} \to \tilde{\mathbf{h}}^\text{E}_t$ and the alignment $a_{st}:=\tilde{\mathbf{h}}^\text{D}_s \cdot \tilde{\mathbf{h}}^\text{E}_t$.

\subsection{Training}

For the one-to-one and eSCAN datasets, we train networks with the ADAM optimizer~\citep{Adam} and an exponentially-decaying learning rate schedule with an initial learning rate of $\eta=0.1$ and a decay rate of $0.9997$ every step (with the exception of the VED networks, in which case we used a decay rate of $0.9999$).
Cross-entropy loss with $\ell_2$ regularization was used and gradients were clipped at a maximum value of $10$.
Both datasets used a batch size of $128$ and each dynamically generated dataset was trained over two epochs.
For the AED and VED architectures, a hidden dimension size of $n=128$ was used, while for the AO we used $n=256$ (for LSTM cells, both the hidden-state $\widetilde{\mathbf{h}}_t$ and the memory $\mathbf{c}_t$ are $n$-dimensional).
For these synthetic experiments, we do not add a bias term to this linear readout layer for the purposes of simplicity and ease of interpretation.
All training for these tasks was performed on GPUs and took at most 20 minutes.

As mentioned above, due to the small vocabulary size of one-to-one and eSCAN, we simply pass one-hot encoded inputs in the RNN architectures,  i.e. we use no embedding layer.
For the AO architecture, the input dimension was padded up $50$ and $100$ for the one-to-one and eSCAN tasks, respectively.
Positional encoding vectors were rotated by a random orthonormal matrix so they were misaligned with the one-hot-encoded input vectors.
Finally, we found performance improved when positional encoding time-scale $\tau$ was chosen to be of order the phrase length: $\tau=50$ for the one-to-one tasks and $\tau=100$ for the eSCAN tasks.

For the natural translation datasets, each token was mapped to an embedded vector of dimension 128 using a learned embedding layer. 
Both the AED and AO architectures used GRU cells, with hidden dimension size of $n=128$. 
As in the synthetic datasets, we train using the ADAM optimizer for 30,000 steps using a batch size of 64; we use an exponential learning rate schedule with initial $\eta=0.01$ and a decay rate of 0.99995.  
Gradients are clipped to a maximum value of $30$.

\section{Additional Results}
\label{app:more_results}

In this section, we discuss several additional results that supplement those discussed in the main text.

\subsection{LSTM and UGRNN Cells}

\begin{figure*}[hp]
    \centering
    \includegraphics[width=0.75\textwidth]{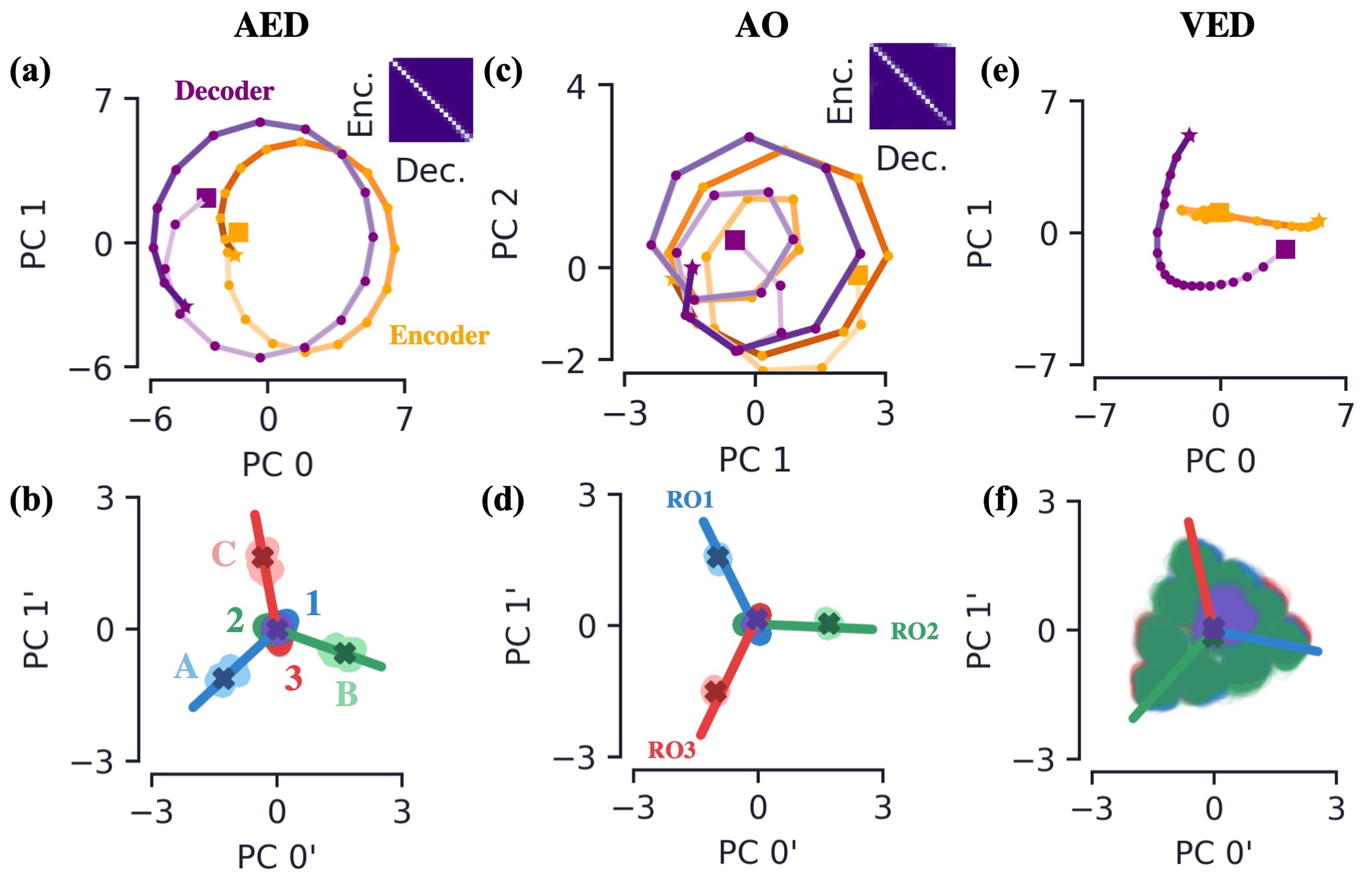}
    \vspace{-3mm}
    \caption{\small
    \textbf{Summary of dynamics for all three architectures on the one-to-one task with UGRNN cells.} 
    All three architectures trained on an $N=3$ one-to-one task of variable length ranging from $15$ to $20$.
    \textbf{(a)} For AED, the path formed by the temporal components of the encoder (orange) and decoder (purple), $\enct{t}$ and $\dect{s}$.
    We denote the first and last temporal component by a square and star, respectively, and the color of said path is lighter for earlier times.
    The inset shows the softmaxed alignment scores for $\dect{s} \cdot \enct{t}$, which we find to be a good approximation to the full alignment for the one-to-one task.
    \textbf{(b)} The input-delta components of the encoder (light) and decoder (dark) colored by word (see labels).
    The encoder input components, $\enci{x}$ are represented by a dark colored `X'.
    The solid lines are the readout vectors (see labels on (d)).
    \textbf{(c, d)} The same plots for the AO network.
    \textbf{(e, f)} The same plots for the VED network (with no attention inset).
    }
    \label{fig:summary_ugrnn}
\end{figure*}

\begin{figure*}[hp]
    \centering
    \includegraphics[width=0.75\textwidth]{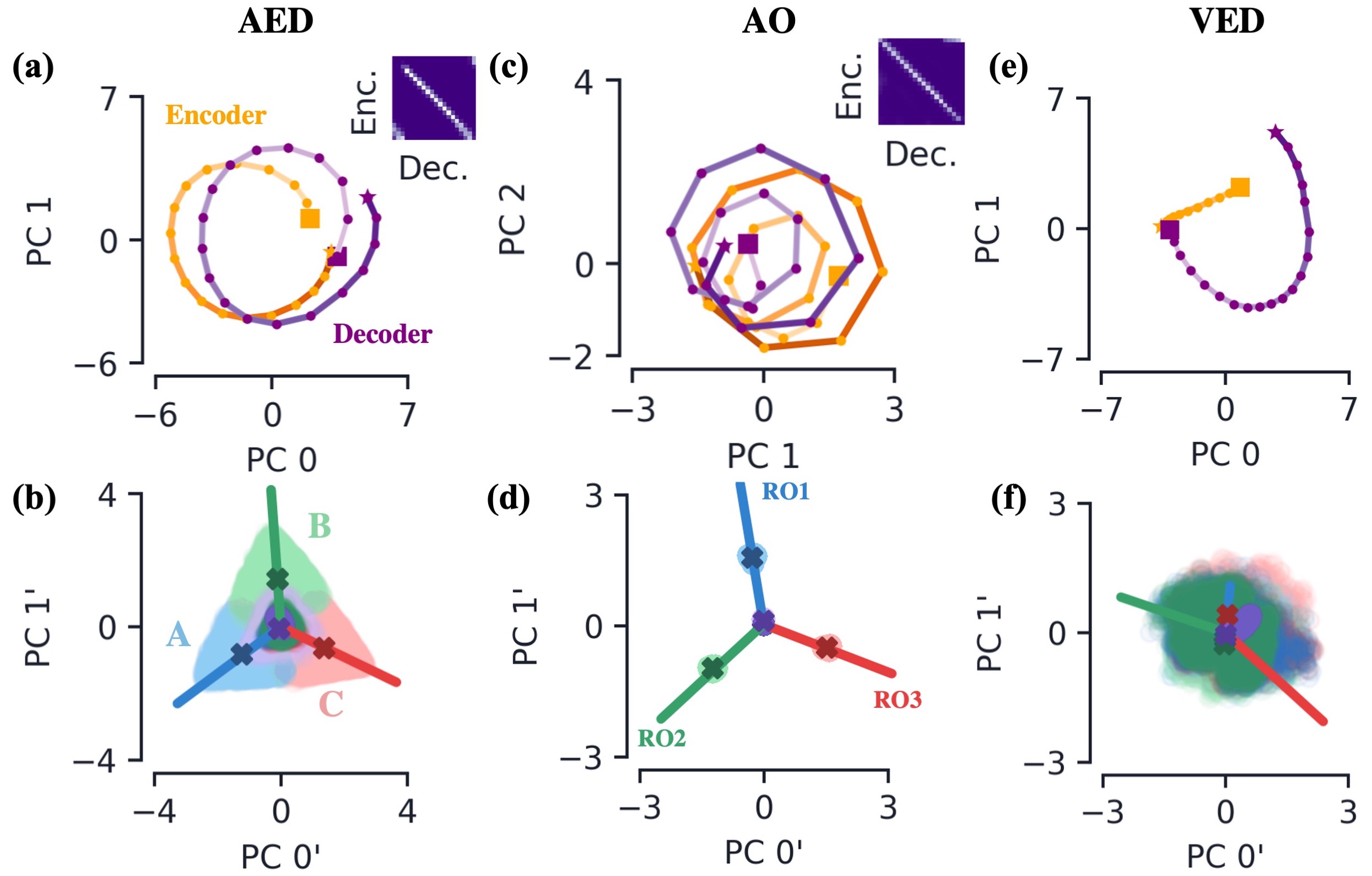}
    \vspace{-3mm}
    \caption{\small
    \textbf{Summary of dynamics for all three architectures on the one-to-one translation task with LSTM cells.} 
    All three architectures trained on an $N=3$ one-to-one translation task with inputs of variable length ranging from $15$ to $20$.
    \textbf{(a)} For AED, the path formed by the temporal components of the encoder (orange) and decoder (purple), $\enct{t}$ and $\dect{s}$.
    We denote the first and last temporal component by a square and star, respectively, and the color of said path is lighter for earlier times.
    The inset shows the softmaxed alignment scores for $\dect{s} \cdot \enct{t}$, which we find to be a good approximation to the full alignment for the one-to-one task.
    \textbf{(b)} The input-delta components of the encoder (light) and decoder (dark) colored by word (see labels).
    The encoder input components, $\enci{x}$ are represented by a dark colored `X'.
    The solid lines are the readout vectors (see labels on (d)).
    \textbf{(c, d)} The same plots for the AO network.
    \textbf{(e, f)} The same plots for the VED network (with no attention inset).
    }
    \label{fig:summary_lstm}
\end{figure*}

In the main text, all plots used a GRU RNN for the cells in the encoder and decoder. 
For the UGRNN and LSTM we find qualitatively similar results on the one-to-one task.
Summaries of our results for the two RNN cells are shown in Fig.~\ref{fig:summary_ugrnn} and Fig.~\ref{fig:summary_lstm}, respectively.
For all types of cells, the networks achieve 100\% test accuracy on the one-to-one task.

Notably, for the AED architecture, we observe that the LSTM has a slightly different attention matrix than that of the GRU and UGRNN. 
Namely, for the first few decoder time steps, the LSTM appears to attend to $\ench{T}$ of a given phrase. 
This means that the network is transferring information about the first few inputs all the way to the final encoder hidden state.
As such, it appears that the AED architecture with LSTM cells relies more on recurrence to solve the one-to-one task relative to its GRU and UGRNN counterparts.
Similar to our findings for the VED architecture in the main text, we find the input-delta components to be significantly less clustered around their corresponding input components when this occurs (Fig.~\ref{fig:summary_lstm}b).
In contrast, the AO architecture with LSTM cells is qualitatively similar to that with the GRU or UGRNN cells. 

\begin{figure*}[hp]
    \centering
    \includegraphics[width=0.9\textwidth]{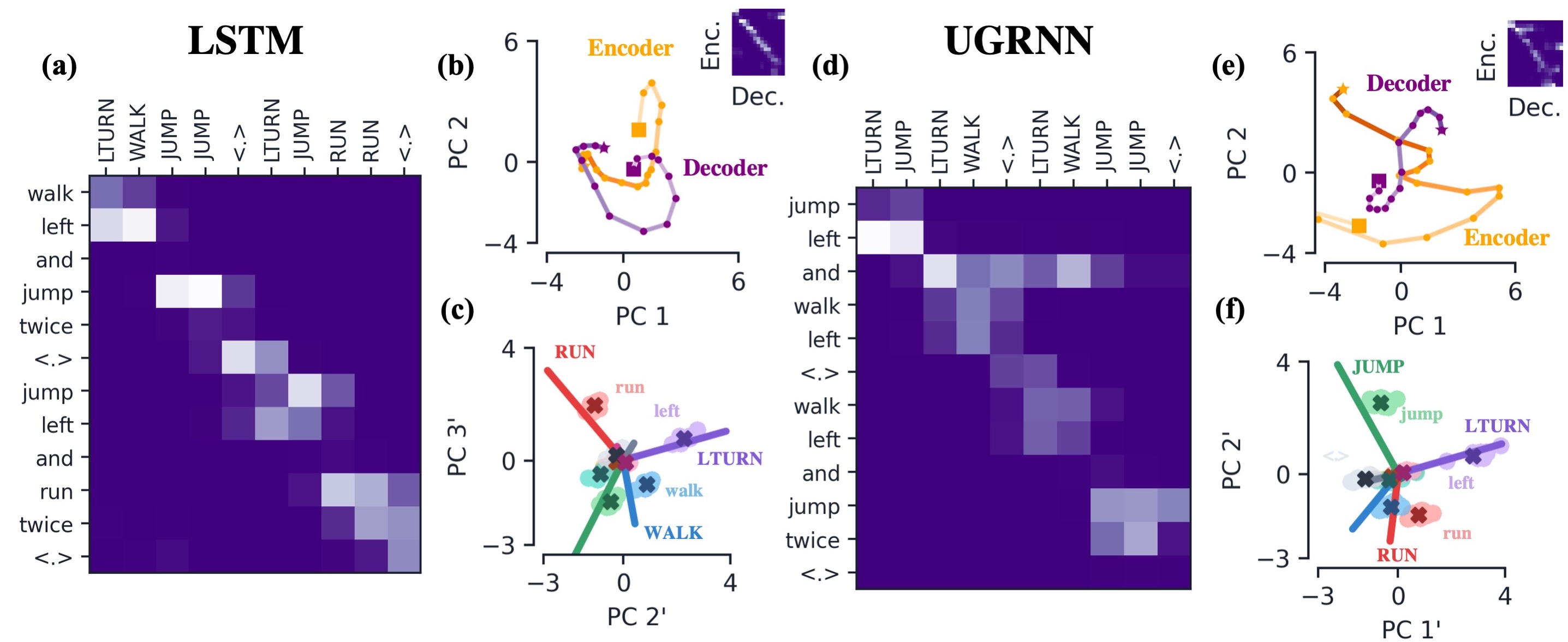}
    \vspace{-3mm}
    \caption{\small
    \textbf{Summary of dynamics for AO architectures with LSTM and UGRNN cells trained on eSCAN.}
    \textbf{(a)} Example attention matrix for the AED architecture.
    \textbf{(b)} AED network's temporal components, with the inset showing the attention matrix from said temporal components. 
    Once again, encoder and decoder components are orange and purple, respectively.  
    \textbf{(c)} AED network's input-delta components, input components, and readouts, all colored by their corresponding input/output words (see labels).
    \textbf{(d, e, f)} The same plots for UGRNN cells.
    }
    \label{fig:ao_lstm_ugrnn}
\end{figure*}

We also train the AO architecture with LSTM and UGRNN cells on eSCAN.
We again see qualitatively similar behavior to what we saw in the main text for AO with GRU cells (Fig.~\ref{fig:ao_lstm_ugrnn}).
For both types of cells, we see the temporal components align to form an approximately diagonal attention matrix (Fig.~\ref{fig:ao_lstm_ugrnn}b,e). 
Once again, we see the input-delta components to be closely clustered around their corresponding input components, and said input components are close to their respective readouts, implementing the translation dictionary (Fig.~\ref{fig:ao_lstm_ugrnn}c,f).

\subsection{Autonomous Dynamics and Temporal Components}

\begin{figure*}[hp]
    \centering
    \includegraphics[width=0.75\textwidth]{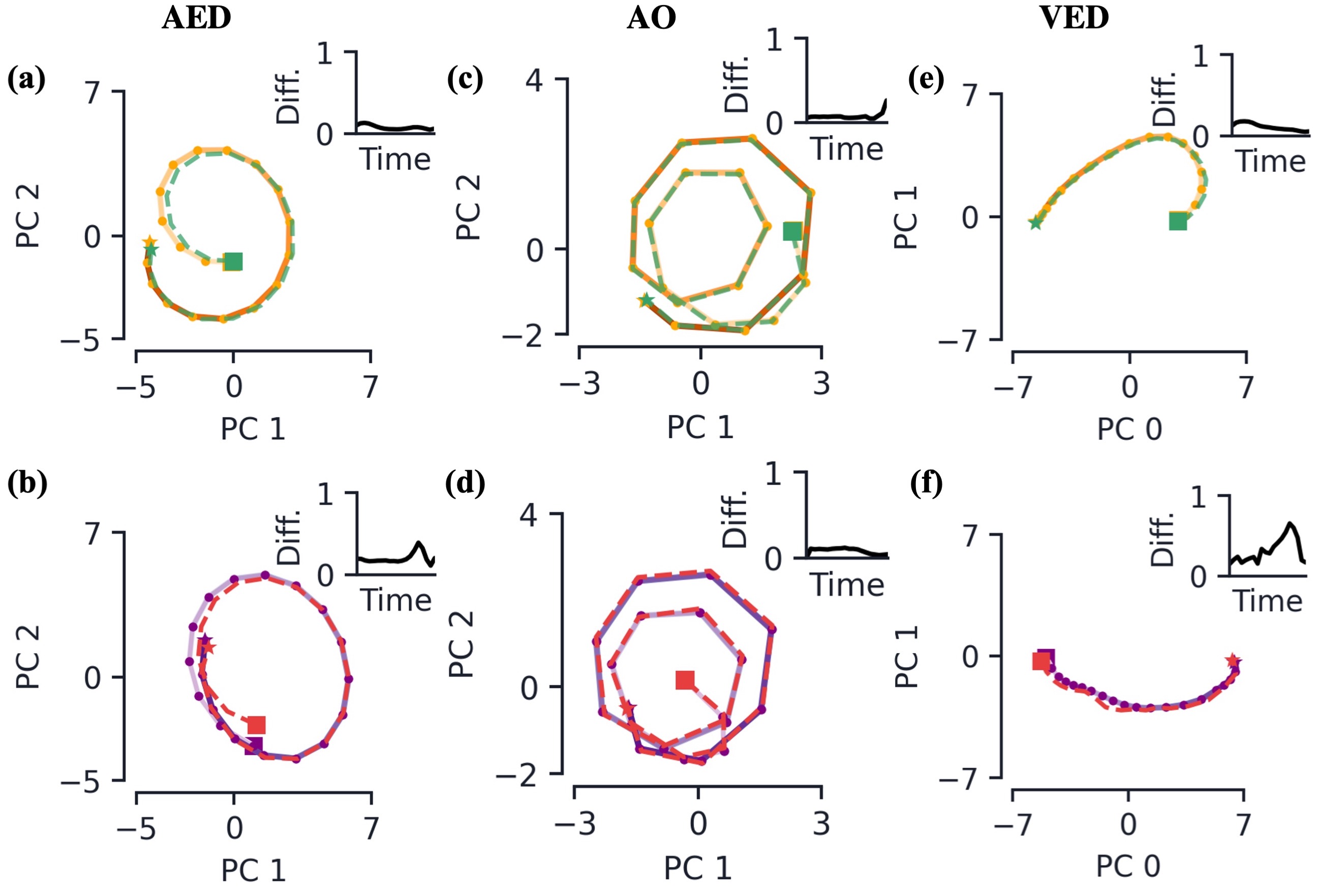}
    \vspace{-3mm}
    \caption{\small
    \textbf{Autonomous dynamics versus temporal components for architectures trained on one-to-one translation.} 
    All three architectures are trained on an $N=3$ one-to-one translation task with inputs of variable length ranging from $15$ to $20$.
    \textbf{(a)} For AED, the path formed by the temporal components of the encoder (orange), $\enct{t}$.
    Also plotted in green are the null hidden states, $\mathbf{h}_t^\text{E,0}$.
    We denote the first and last hidden state of each of these by a square and star, respectively. 
    The inset shows the quantitative difference between the two states, $\|\ench{t} - \mathbf{h}_t^\text{E,0} \|_2/\| \ench{t}\|_2$, as a function of encoder time step, $t$. 
    \textbf{(b)} The same plot but for the decoder temporal components (purple) and the decoder null hidden states (red).
    \textbf{(c, d)} The same plots for the AO network.
    \textbf{(e, f)} The same plots for the VED network.
    }
    \label{fig:auto_oto}
\end{figure*}

When the temporal components are the leading-order behavior in alignment scores (e.g. in the one-to-one and eSCAN tasks), we find them to be well approximated the hidden states resulting from the network having zero input, i.e. $\mathbf{x}_t^\text{E}=\mathbf{0}$ for all $t$. 
In the case of AED and VED, this means the encoder RNNs are driven solely by their recurrent behavior.
For AO, the encoder RNNs are driven by only the positional encoding vectors.
Since in all three cases this results network outputting hidden states independent of the details of the input, we call this the \emph{autonomous} dynamics of the network. 

Denote the encoder and decoder hidden states resulting from no input by $\mathbf{h}_t^\text{E,0}$ and $\mathbf{h}_s^\text{D,0}$, respectively.
For AED and VED, they are
\begin{align}
    \mathbf{h}_t^\text{E,0} = F_\text{E}(\mathbf{h}_{t-1}^\text{E,0}, \mathbf{0})\,, \qquad 
    \mathbf{h}_s^\text{D,0} = F_\text{D}(\mathbf{h}_{s-1}^\text{D,0},\mathbf{0})\, .
\end{align}
For the AO network, we still add the positional encoding vectors to the inputs,
\begin{align}
    \mathbf{h}_t^\text{E,0} = F_\text{E}(\mathbf{0}, \mathbf{p}_t^\text{E})\, , \qquad 
    \mathbf{h}_s^\text{D,0} = F_\text{D}(\mathbf{0},\mathbf{p}_s^\text{D}) \,. 
\end{align}
Plotting the resulting hidden states along with the temporal components, we find for all three architectures the two quantities are quite close at all time steps of the encoder and decoder (Fig.~\ref{fig:auto_oto}).
We quantify the degree to which the null hidden states approximate the temporal components via $\|\ench{t} - \mathbf{h}_t^\text{E,0}\|_2/\| \ench{t}\|_2$ and $\|\dech{s} - \mathbf{h}_s^\text{D,0} \|_2/\| \dech{s}\|_2$ where $\left\|\cdot\right\|_2$ denotes the $\ell_2$-norm. 
For the AO network, we find the average of this quantity across the entire encoder and decoder phrases to be about $0.07$ and $0.08$, respectively, while for AED we find it to be $0.07$ for the encoder and $0.19$ for the decoder.

\begin{figure}[ht]
    \centering
    \includegraphics[width=0.7\textwidth]{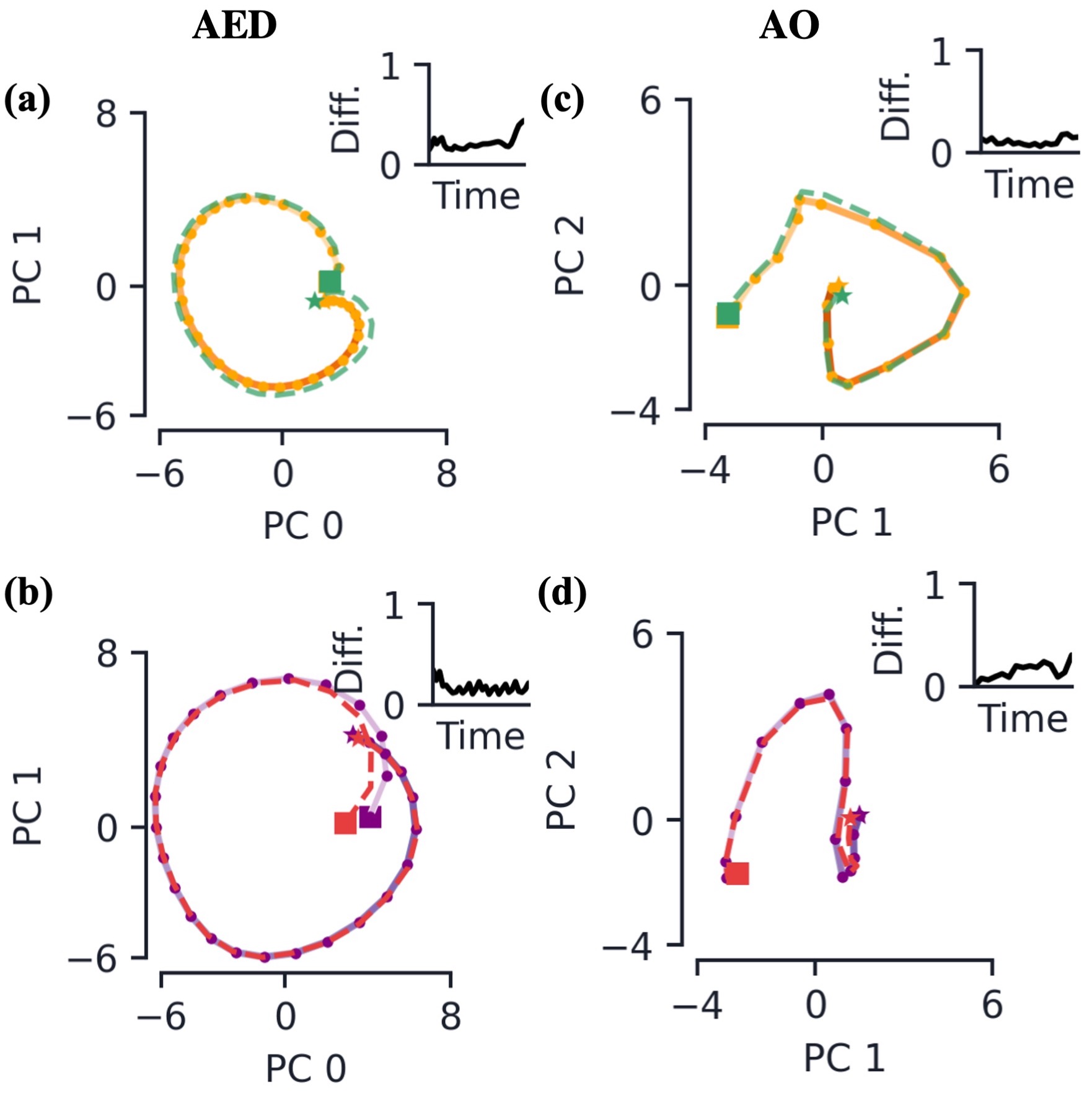}
    \vspace{-1mm}
    \caption{\small
    \textbf{Autonomous dynamics versus temporal components for architectures trained on eSCAN.}
    \textbf{(a)} For AED, the path formed by the temporal components of the encoder (orange), $\enct{t}$.
    Also plotted in green are the null hidden states, $\mathbf{h}_t^\text{E,0}$.
    We denote the first and last hidden state of each of these by a square and star, respectively. 
    The inset shows the quantitative difference between the two states, $\|\ench{t} - \mathbf{h}_t^\text{E,0} \|_2/\| \ench{t}\|_2$, as a function of encoder time step, $t$. 
    \textbf{(b)} The same plot but for the decoder temporal components (purple) and the decoder null hidden states (red).
    \textbf{(c, d)} The same plots for the AO network.
    }
    \label{fig:auto_scan}
\end{figure}

We also find the null hidden states to be close to the temporal components of eSCAN (Fig.~\ref{fig:auto_scan}).
Again, averaging our difference measure across the entire encoder and decoder phrases, for AO we find $0.11$ and $0.15$ and for AED $0.21$ and $0.17$, respectively.

This result gives insight into the network dynamics that drive the behavior of the temporal components.
In AED and VED, each RNN cell is driven by two factors: the recurrent hidden state and the input. The AO architecture is similar, but the recurrence is replaced by temporal information through the positional encoding.
The absence of input eliminates the input-driven behavior in the RNN cells.
Since the network's hidden states still trace out paths very close to the temporal components, this is evidence that it is the recurrent dynamics (in the case of AED and VED) or the positional encoding vectors that drive the network's temporal component behavior. 

One can use this information to postulate other behaviors in the network. 
For instance, given the lack of correlation between inputs in the one-to-one translation task, in the AO network one may wonder how much of the decoder's dynamics are driven by recurrent versus input behavior.
We already know the decoder's primary job in this network is to align its temporal components with that of the encoder, and the results above suggest said behavior is driven primarily by the positional encoding vectors and not the inputs.
To test this, we compare the accuracies of a trained network with and without inputs into the decoder RNN. 
We still achieve 100\% word accuracy when $\mathbf{x}^\text{D}_s = \mathbf{0}$ for all decoder time steps.

\subsection{Learned Attention}
\label{app:learned_attention}

In addition to the dot-product attention considered throughout the main text, we have implemented and analyzed new architectures that are identical to the AED and AO architectures that use a learned-attention mechanism. 
These networks use a scaled-dot product attention in the form of queries, keys, and value matrices similar to the original Transformer \cite{vaswani2017attention}.
More specifically, the context vector $\mathbf{c}_s$ and alignment $a_{st}$ are now determined by the expressions
\begin{align}
    \mathbf{c}_s := \sum_{t=1}^T \alpha_{st} \mathbf{v}_t \,, \qquad a_{st}:= \mathbf{q}_{s} \cdot \mathbf{v}_{t}\,. 
\end{align}
In these expressions, the vectors $\mathbf{v}_t$, $\mathbf{q}_s$, and $\mathbf{k}_t$ are product of the \emph{learned} weight matrices $\mathbf{V} \in \mathbb{R}^{n\times n}$, $\mathbf{Q} \in \mathbb{R}^{n^\prime\times n}$, and $\mathbf{K} \in \mathbb{R}^{n^\prime\times n}$ and the hidden states,
\begin{align}
    \mathbf{v}_t := \mathbf{V} \ench{t}\,, \qquad
    \mathbf{q}_s := \mathbf{Q} \dech{s}\,, \qquad
    \mathbf{k}_t := \mathbf{K} \ench{t}\,,
\end{align}
with $\mathbf{v}_t \in \mathbb{R}^n$ (i.e. the same dimension as hidden state space) and $\mathbf{q}_s, \mathbf{k}_t \in \mathbb{R}^{n^\prime}$, with $n^\prime$ the dimension of the query/key space. 

\begin{figure}[t]
    \centering
    \includegraphics[width=0.7\textwidth]{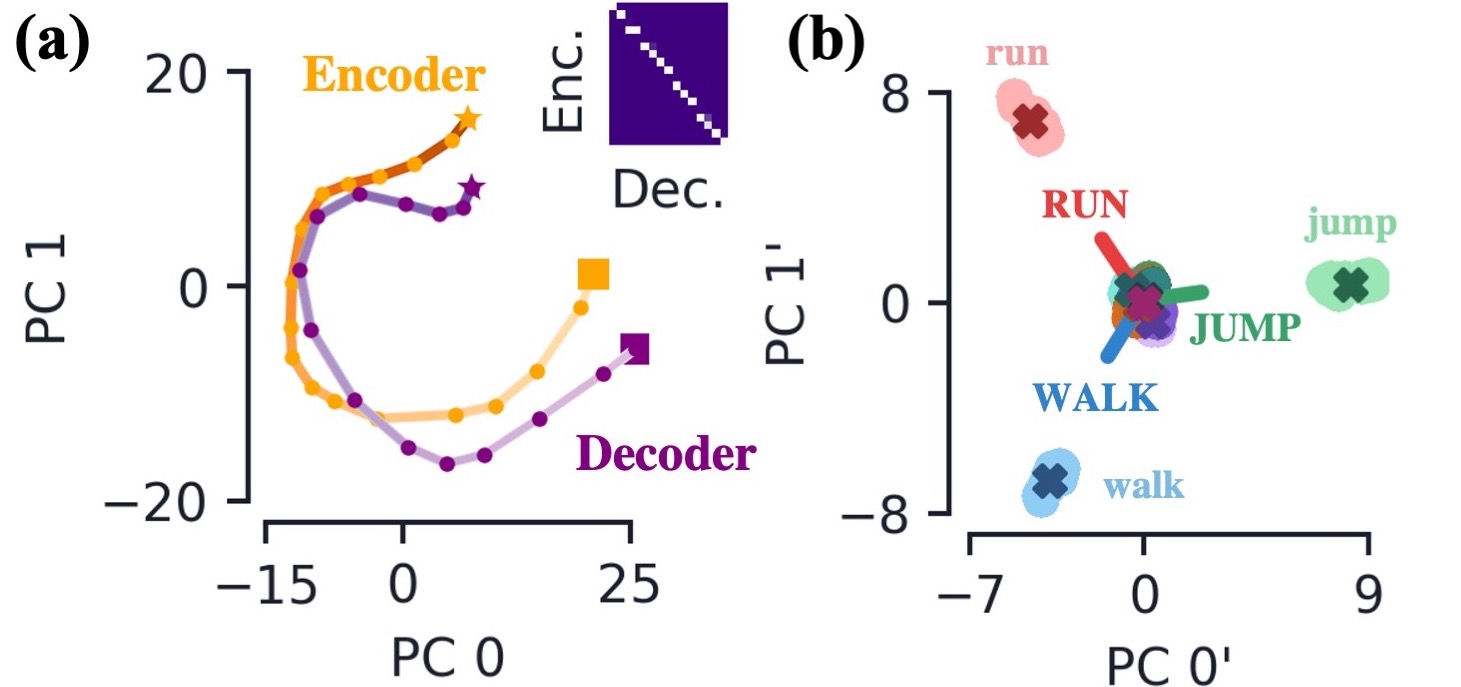}
    \vspace{-1mm}
    \caption{\small
    \textbf{Dynamics for AO with learned attention.} 
    \textbf{(a)} The path formed by the temporal components of the encoder (orange) and decoder (purple) multiplied by the key and query matrices, respectively, i.e. $\mathbf{K}\enct{t}$ and $\mathbf{Q}\dect{s}$.
    We denote the first and last temporal component by a square and star, respectively, and the color of said path is lighter for earlier times.
    The inset shows the softmaxed alignment scores for $\mathbf{q}_{s} \cdot \mathbf{k}_{t}$, which we find to be a good approximation to the full alignment for the one-to-one translation task.
    \textbf{(b)} The input-delta components of the encoder (light) and decoder (dark) colored by word (see labels), after being multiplied by value matrix, $\mathbf{V}$.
    The encoder input components, $\mathbf{V}\enci{x}$ are represented by a dark colored `X'.
    The solid lines are the readout vectors.
    }
    \label{fig:learned}
\end{figure}

After training these networks on our one-to-one and eSCAN tasks, we find very similar results to that of dot-product attention. 
In particular, temporal components of the encoder and decoder continue to align with one another after being projected through the respective key/query matrix (for AO, see Fig.~\ref{fig:learned}a). 
Input-delta components cluster along the respective readouts, after being projected through the value matrix (Fig.~\ref{fig:learned}b). 
Decomposing the alignment scores for these networks, we continue to find them to be dominated by the temporal component term (i.e. $\dect{s} \mathbf{Q}^T \mathbf{K} \enct{t}$).

This gives us additional confidence that the analysis techniques can be applied to modern architectures that use attention. For instance, in a multi-headed attention setting, such a decomposition could be used on each head separately to characterize the dynamics behind each of the heads. 

\subsection{Attention Only with Non-Gated Feedforward}

\begin{figure}[h]
    \centering
    \includegraphics[width=0.7\textwidth]{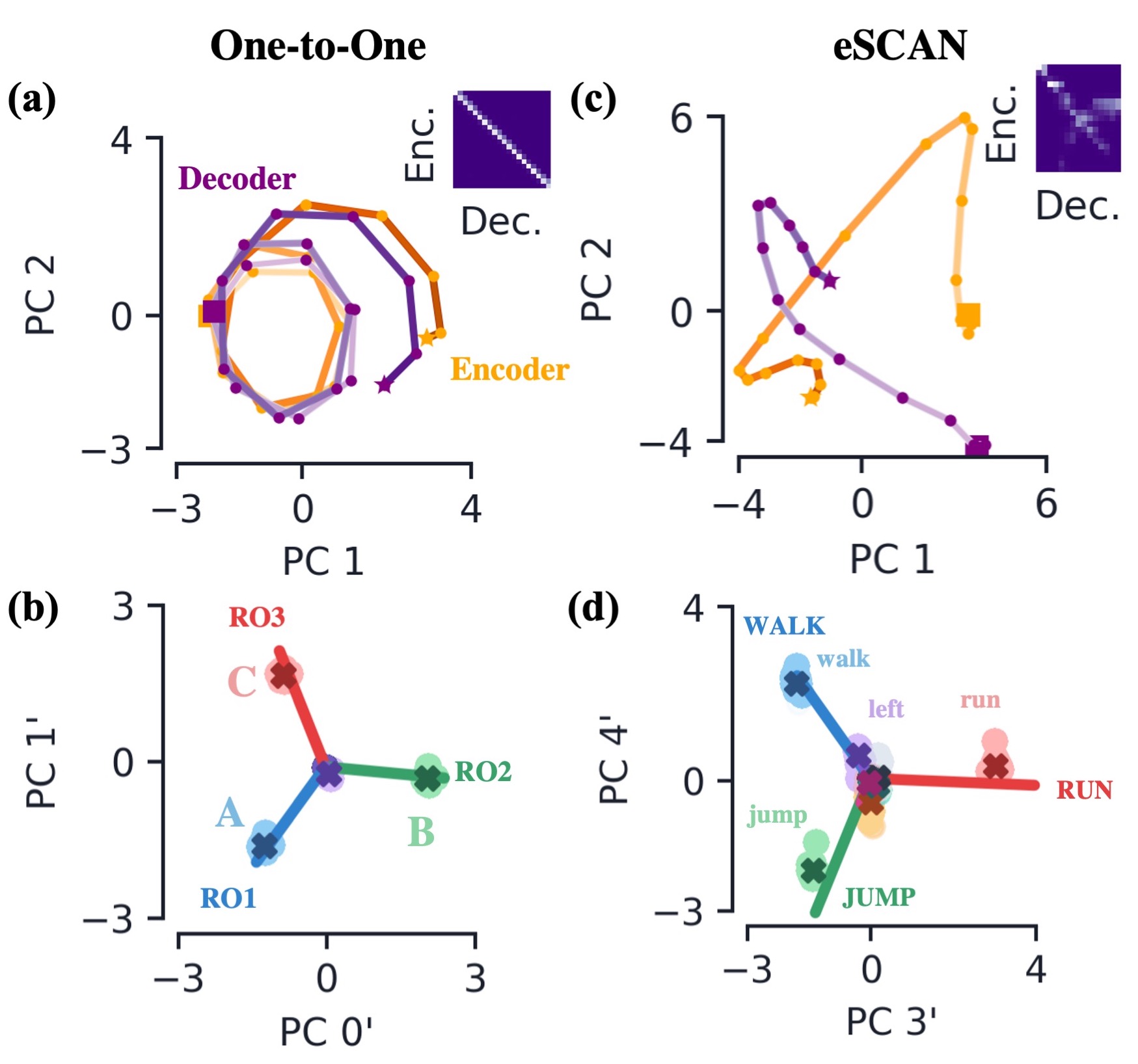}
    \vspace{-1mm}
    \caption{\small
    \textbf{AO with non-gated feed foward trained on one-to-one and eSCAN.} 
    \textbf{(a)} For the AO trained on an $N=3$ one-to-one translation task, the path formed by the temporal components of the encoder (orange) and decoder (purple), $\enct{t}$ and $\dect{s}$.
    We denote the first and last temporal component by a square and star, respectively, and the color of said path is lighter for earlier times.
    The inset shows the softmaxed alignment scores for $\dect{s} \cdot \enct{t}$, which we find to be a good approximation to the full alignment for the one-to-one translation task.
    \textbf{(b)} The input-delta components of the encoder (light) and decoder (dark) colored by word (see labels).
    The encoder input components, $\enci{x}$ are represented by a dark colored `X'.
    The solid lines are the readout vectors (see labels).
    \textbf{(c, d)} The same plots for AO trained on eSCAN.
    }
    \label{fig:non_gated}
\end{figure}

The AO architecture's feedfoward networks are created by zeroing the recurrent part of various RNNs. 
Although this does result in a feedforward network, the presence of the gating mechanisms in the RNN cells we use in this work make this non-standard feedforward network.
To verify our qualitative results hold beyond a gated feedfoward network, we investigated if our results differed when using a standard fully connected layer followed by a tanh readout.

We find the non-gated AO network trains well on both the one-to-one and eSCAN tasks, achieving 100\% and 98.8\% word accuracy, respectively.
Again performing the temporal and input component decomposition on the network's hidden states, we find the qualitative dynamics of this network are the same as its RNN counterparts (Fig.~\ref{fig:non_gated}).
For example, in the one-to-one translation task we find the temporal components of the encoder and decoder again mirror one another in order to from a diagonal attention matrix (Fig.~\ref{fig:non_gated}a).
The input components of the encoder align with the readouts to implement the translation dictionary (Fig.~\ref{fig:non_gated}b).

\subsection{Encoder-Decoder with Attention Readouts}

\begin{figure}[h]
    \centering
    \includegraphics[width=0.7\textwidth]{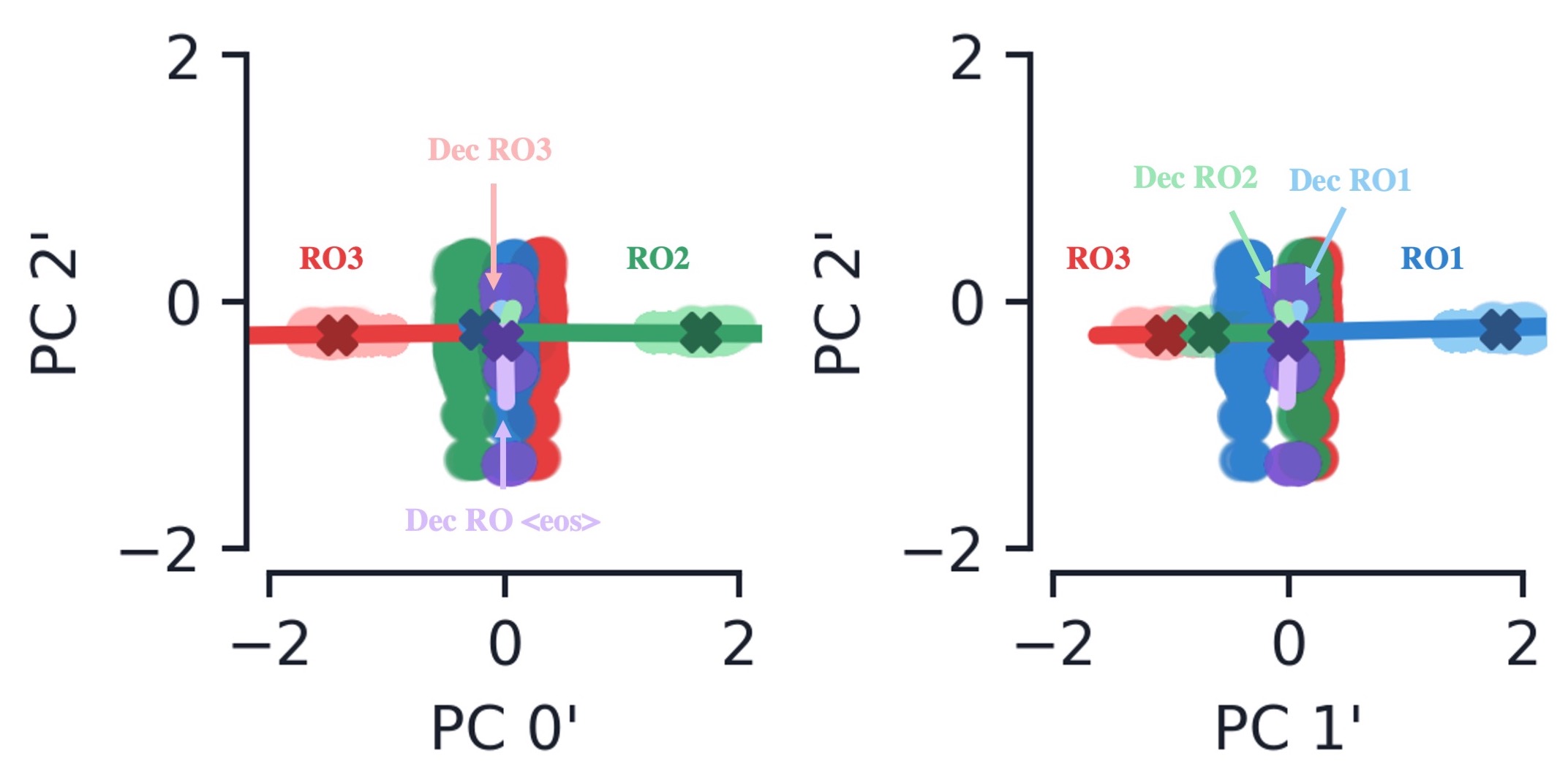}
    \vspace{-1mm}
    \caption{\small
    \textbf{Decoder hidden state behavior in AED trained on one-to-one translation.} 
    The input-delta components of the encoder (light) and decoder (dark) colored by word.
    The encoder input components, $\enci{x}$ are represented by a dark colored `X'.
    The solid lines are the readout vectors, with those corresponding to the context vector colored dark (and labeled by `RO') and those corresponding to the decoder hidden state readout colored light (and labeled by `Dec RO').
    }
    \label{fig:dec_readouts}
\end{figure}

The AED network's linear readout takes into account both the decoder hidden state output and the context vector, i.e. $\mathbf{y}_s = \mathbf{W}[\dech{s}, \mathbf{c}_s]$.
As such, each output's readout can be viewed as \emph{two} vectors in hidden state space: one which acts on the context vector and the other which acts on the decoder hidden state output.
Here we elaborate on our comment in the main text regarding the effects of the decoder readouts being negligible.

Recall that after training the AED network, we found the context vector readouts align with the encoder input components, yielding the translation dictionary for a given task.
We find the the decoder readouts to be close to orthogonal to the context vector readouts (Fig.~\ref{fig:dec_readouts}).
Furthermore, we find the readouts for all words other than the `eos' character to be closely aligned.
Omitting the `eos` effect, this results in the decoder readouts contributing roughly equal values to all logit values.
Since the logit values are passed through a softmax function, it is their differences that ultimately matter when it comes to classifying a given output as a word.
In contrast, we found the context vector readouts to align with the vertices of an $(N-1)$-simplex and the logit value contributions to vary significantly more with the hidden state. 
Indeed, comparing the difference between the largest and second largest logit contribution of each set of readouts, we find the difference due to the context vector readouts to be several times larger than that of the decoder readouts.
For AED trained on eSCAN, we again find the differences in the context vector logit contributions to be several times larger than that due to the decoder hidden states.

\subsection{Vanilla Encoder-Decoder Dynamics}
\label{app:ved_dyn}

\begin{figure}[h]
    \centering
    \includegraphics[width=0.7\textwidth]{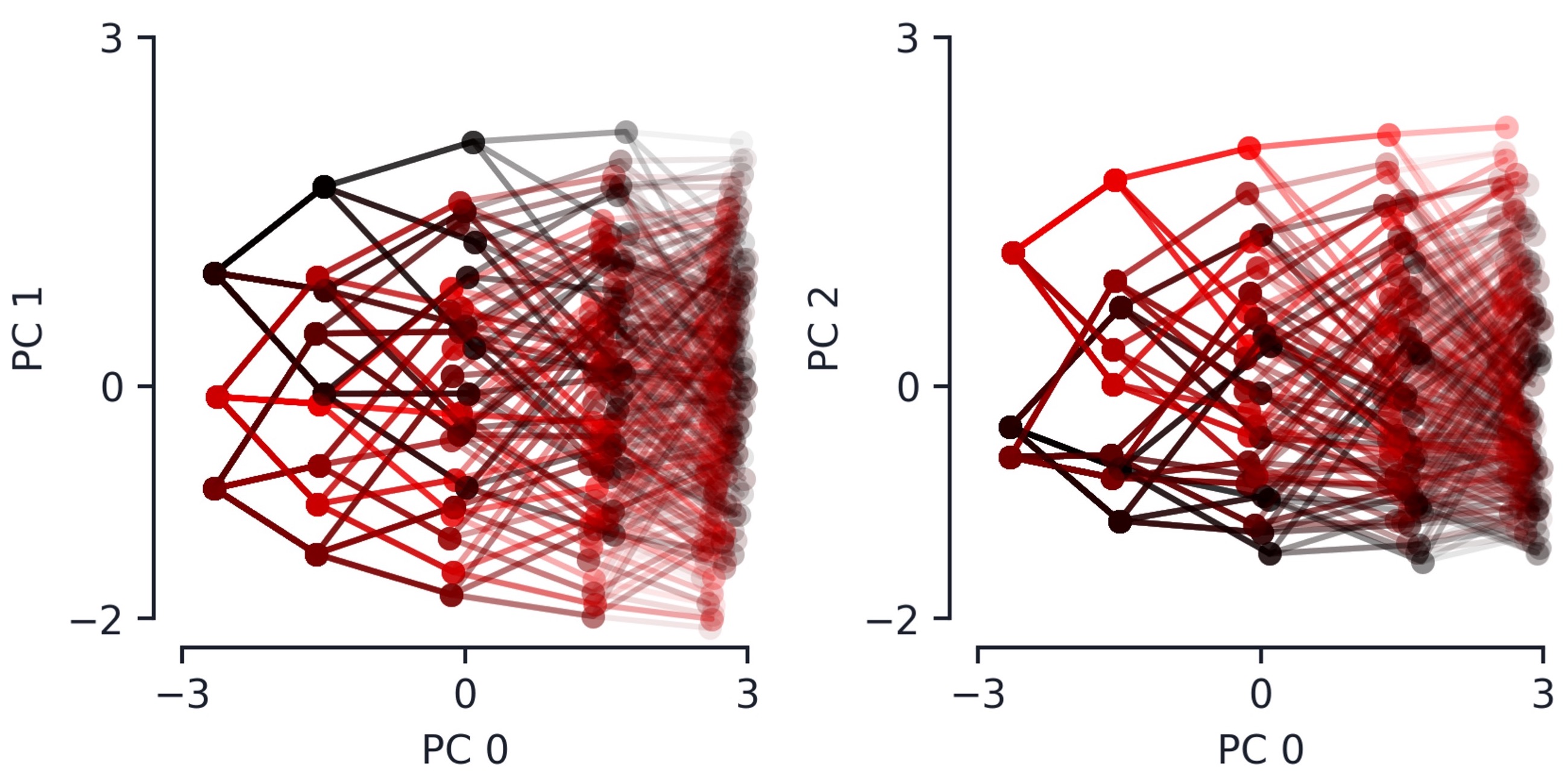}
    \vspace{-1mm}
    \caption{\small
    \textbf{Tree-like structure of encoder hidden states in VED.} 
    The encoder hidden states of the VED architecture trained on the one-to-one translation task for the first five time steps.
    Note the hidden states organize themselves in five distinct clusters along PC 0.
    Also shown are the paths in hidden state space various input phrases take as a given input phrase is read.
    Said paths are colored from red to black by similarity starting from latest time, i.e. $\{\text{B}, \text{A}, \text{C}, \text{A}, \text{C}\}$ and $\{\text{B}, \text{A}, \text{C}, \text{A}, \text{B}\}$ are similarly colored but $\{\text{C}, \text{A}, \text{C}, \text{A}, \text{C}\}$ is not.
    }
    \label{fig:ved_tree}
\end{figure}

In the main text, we briefly discussed the dynamics of the VED architecture, and here we provide some additional details.
After training the VED architecture on the one-to-one translation task, we found that the encoder and decoder hidden states belonging to the same time step formed clusters, and said clusters are closest to those corresponding to adjacent time steps.
Additionally, since the VED arhcitecture has no attention, the encoder and decoder hidden states have to carry all relevant information from preceding steps.
To facilitate this, the dynamics of the encoder's hidden state space organizes itself into a tree structure to encode the input phrases (Fig.~\ref{fig:ved_tree}). 
Starting from the encoder's initial hidden state, the hidden states of a given input phrase traverse the branches of said tree as the phrase is read in, ultimately arriving at one of the tree's leaves at time $T$. 
The distinct leaves represent the different final encoder hidden states, $\ench{T}$, that must encode the input phrase's words and their ordering. 

Since the decoder receives no additional input, the encoder must place $\ench{T}$ in a location of hidden state space for the decoder's dynamics to produce the entire output phrase.
Although the network does indeed learn to do this, we do not observe the reversal of the tree structure learned by the encoder.
That is, any two phrases that have the same output sequence for any time $s \geq s^\prime$ could occupy the same decoder hidden states $\tilde{\mathbf{h}}_s$ for $s \geq s^\prime$. 
This would result in the temporal mirror of the tree structure seen in the encoder. 
However, such a structure is not observed and the decoder instead seems to arrive at a solution where all output paths are kept distinct.

\end{document}